\documentclass{article}

\usepackage[preprint]{corl_2024} 
\usepackage{amssymb}            
\usepackage{mathtools}          
\usepackage{mathrsfs}   

\usepackage{graphicx}           
\usepackage{subcaption}         
\usepackage[space]{grffile}     
\usepackage{url}                
\usepackage{booktabs} 
\usepackage{wrapfig}
\usepackage{algorithm}
\usepackage{algpseudocode}
\usepackage{multicol}
\usepackage{makecell}
\usepackage[bottom]{footmisc} 

\newcommand{\ourmethod}{\texttt{$\varepsilon$-retrain}}

\newtheorem{definition}{Definition}

\newtheorem{lemma}{Lemma}
\newtheorem{corollary}{Corollary}

\title{Improving Policy Optimization via $\varepsilon$-Retrain}

\author{
  Luca Marzari \thanks{Work performed while at Carnegie Mellon University (CMU)}\\
  University of Verona \\
  Verona, Italy   \\
  \texttt{luca.marzari@univr.it} \\
  \And
   Priya L. Donti \\
  Massachusetts Institute of Technology \\
  Boston, USA \\
  \texttt{donti@mit.edu} \\
  \AND
  Changliu Liu \\
  Carnegie Mellon University \\
  Pittsburgh, USA \\
  \texttt{cliu6@andrew.cmu.edu} \\
  \And
  Enrico Marchesini \\
  Massachusetts Institute of Technology\\
  Boston, USA \\
  \texttt{emarche@mit.edu} \\
}

\begin{document}
\maketitle


\begin{abstract}
We present \ourmethod, an exploration strategy encouraging a behavioral preference while optimizing policies with monotonic improvement guarantees. To this end, we introduce an iterative procedure for collecting \textit{retrain areas}---parts of the state space where an agent did not satisfy the behavioral preference. Our method switches between the typical uniform restart state distribution and the retrain areas using a decaying factor $\varepsilon$, allowing agents to retrain on situations where they violated the preference. We also employ formal verification of neural networks to provably quantify the degree to which agents adhere to these behavioral preferences. Experiments over hundreds of seeds across locomotion, power network, and navigation tasks show that our method yields agents that exhibit significant performance and sample efficiency improvements. 
\end{abstract}



\section{Introduction}

By balancing the trade-off between \textit{exploration} and \textit{exploitation}, a reinforcement learning (RL) agent typically relies on a scalar reward function to learn behaviors capable of solving a task \citep{silver2021reward}. However, these functions often lead to unforeseen behaviors, making it difficult to enforce particular behaviors that we desire the system to exhibit \citep{amodei2016rewards}---a \textit{behavioral preference}.  

For example, consider applying a policy optimization RL method to a robot learning to reach random targets. Commonly, the agent gets a positive reward based on its distance from the goal, a penalty for collisions \citep{drl_navigation1, drl_navigation2}, and we use a uniform restart distribution throughout the state space to randomly initialize the environment at each episode. In this setup, learning good navigation behaviors while satisfying a behavioral preference (or desiderata, interchangeably) related to safety such as \textit{``avoid collisions''} requires many collisions around the same state. However, the uniform restart distribution naturally makes it harder for agents to experience these similar collisions over time despite being pivotal for guaranteeing monotonic policy improvement \citep{kakade2002approximately}. This potentially translates into a higher variance in the local estimate of the objective \citep{TRPO}, making it hard to effectively enforce the desired behavior \citep{eysenbach2018reset, jian2023generalization}.

Previous policy-based approaches investigate the impact of restoring the environment to specific states to improve performance while maintaining theoretical guarantees on monotonic improvement \citep{lagoudakis2003reinforcement,gabillon2013approximate,VFS}. 
A leading example is the \textit{vine} Trust Region Policy Optimization (TRPO) algorithm \citep{TRPO}, designed to enhance exploration and reduce the variance of gradient updates. TRPO \textit{vine} restarts the agent in states visited by the current policy to generate additional rollouts from that state and reduce the policy update variance.
The authors demonstrate how the theoretically justified procedure retains monotonic policy improvement guarantees. 
However, this method and, more generally, designing a poor restart state distribution has three critical downsides we address in our work. 
\begin{itemize}
    \item To the best of our knowledge, policy optimization works have not considered restarting distribution mechanisms geared towards improving specific behavioral preferences.
    \item Sub-optimal distributions can cause approximate methods to get stuck in local optima, potentially resulting in poor agent performance \citep{kakade2002approximately, lagoudakis2003reinforcement}.
    \item \textit{vine} significantly hinders sample efficiency, requiring many additional rollouts for each policy update.
\end{itemize}

\begin{figure}[t]
    \centering
    \includegraphics[width=0.8\linewidth]{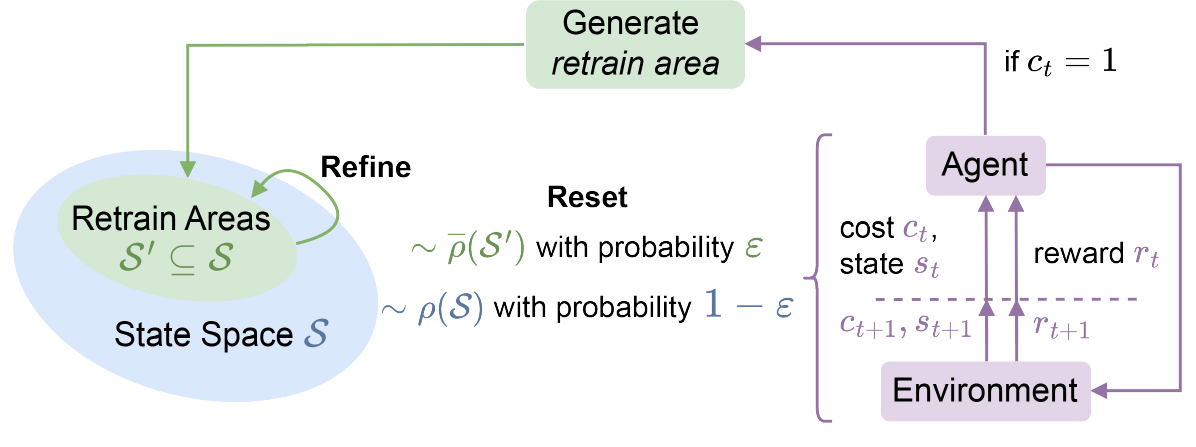}
    \vspace{-5pt}
    \caption{Explanatory overview of \ourmethod.}
    \label{fig:overview}
    \vspace{-8pt}
\end{figure}

This paper presents \ourmethod, a novel exploration strategy designed to optimize policies, maintaining monotonic improvement guarantees while encouraging a behavioral preference. Our method is inspired by human learning, where consistently repeating tasks enhances the learning of a particular behavior. As detailed in Figure \ref{fig:overview}, we exploit an $\varepsilon$ decay strategy to combine the uniform restart state distribution over the state space (blue), typical of RL algorithms, with a restart strategy over \textit{retrain areas} (green). The latter uses an iterative procedure that collects and refines (i.e., creates and merges) portions of the state space where the agent violates the behavioral preference at training time. The proposed approach ``retrains" the agent from these areas, improving the advantage estimation of actions violating the desired behavior, according to a probability $\varepsilon$ (purple). A decaying schedule for $\varepsilon$ also allows us to maintain the asymptotic convergence properties of the underlying RL algorithm and the design of retraining areas avoids sub-optimal distributions as demonstrated by our experiments. Our method also does not require additional rollouts and we prove that using mixed uniform restart distributions leads, in the worst case, to the same monotonic improvement guarantees as in \citet{TRPO}.

We first show the benefits of employing \ourmethod~in adhering to behavioral preferences in an unconstrained setup (where we penalize the reward upon violating the preference). To this end, we evaluate \ourmethod~on top of policy optimization methods (i.e., TRPO \citep{TRPO} and Proximal Policy Optimization (PPO) \citep{PPO}) over hundreds of seeds and different behavioral preferences related to safety---velocity limits in locomotion tasks, preventing overloads in power grids, and collision avoidance in mobile navigation. Following our interest in safe behaviors, we also combine \ourmethod~with the Lagrangian implementations of TRPO and PPO since they have been recently employed to enforce similar behaviors \citep{julien2022behavioralspec, SafetyGymnasium, stooke2020lagpid}. These Lagrangian algorithms are widely used in safe RL literature, and we use them as additional baselines for a more comprehensive evaluation. 
Our experiments consider diversified tasks ranging from simulated locomotion to optimizing power networks and robotic navigation, which is a commonly employed task in the RL literature \citep{drl_navigation1, EPS, drl_navigation2, marl_navigation}. The results show that enhancing policy optimization methods with \ourmethod~leads to significantly higher sample efficiency and better enforce the desiderata while solving the tasks. Additionally, we note that system designers typically evaluate agents empirically and can not provably quantify the degree to which they adhere to the behavioral preferences. Since our theory refers to the improvement over the main reward objective, we employ a formal verification (FV) of neural networks tool to provably quantify the rate at which the agent trained for the navigation task avoids collisions in the retrain areas.\footnote{We use navigation as an explanatory task for clarity since it allows us to visualize retrain areas.} Finally, the realistic environment employed in the navigation task
enables the transfer of policies trained in simulation on ROS-enabled
platforms. Hence, we show the effectiveness of \ourmethod\;in a realistic unsafe navigation scenario.

\section{Preliminaries and Related Work}\label{preliminaries}

We consider problems defined as Markov decision processes (MDPs), modeled as a tuple $(\mathcal{S}, \mathcal{A}, \mathcal{P}, \rho, R, \gamma)$; $\mathcal{S}$ and $\mathcal{A}$ are the finite sets of states and actions, respectively, $\mathcal{P}:\mathcal{S \times \mathcal{A} \times \mathcal{S}} \to [0, 1]$ is the state transition probability distribution, $\rho: \mathcal{S}\to [0, 1]$ is the initial uniform state distribution, $R: \mathcal{S}\times\mathcal{A} \to \mathbb{R}$ is a reward function, and $\gamma\in [0,1)$ is the discount factor. In policy optimization algorithms, agents learn a parameterized stochastic policy $\pi:\mathcal{S} \times \mathcal{A} \to [0,1]$, modeling the probability to take an action $a_t \in \mathcal{A}$ in a state $s_t \in \mathcal{S}$ at a certain step $t$. The goal is to find the parameters that maximize the expected discounted reward $\psi(\pi) = \mathbb{E}_{\tau\sim\pi} [\sum_{t=0}^\infty \gamma^t R(s_t, a_t)]$,
where $\tau:= (s_0, a_0, s_1, a_1, \dots)$ is a trajectory with $s_0 \sim \rho(s_0)$, $a_t \sim \pi(a_t \vert s_t)$, $s_{t+1} \sim \mathcal{P}(s_{t+1} \vert s_t, a_t)$. We also define state and action value functions $V_\pi$ and $Q_\pi$ modeling the expected discount return starting from the state $s_t$ (and action $a_t$ for $Q_\pi$) and following the policy $\pi$ thereafter as: $V_\pi(s_t) = \mathbb{E}_{a_t, s_{t+1}, a_{t+1},\dots}[ \sum_{i=0}^\infty \gamma^i R(s_{t+i}, a_{t+i})]$ and $Q_\pi(s_t, a_t) = \mathbb{E}_{s_{t+1}, a_{t+1},\dots}[ \sum_{i=0}^\infty \gamma^i R(s_{t+i}, a_{t+i})]$.
Given the current state and action, we can also measure how much better or worse the agent performs compared to its expected performance---the advantage function $A_\pi(s,a) = Q_\pi(s,a) - V_\pi(s)$.

To derive a bound on the policy improvement, \citet{TRPO} also define the expected advantage of a new policy $\pi'$ over the old $\pi$, and relate the expected discounted return of $\pi'$ to $\pi$: $\Tilde{A}(s) = \mathbb{E}_{a \sim \pi'(\cdot \vert s)}[A_\pi(s,a)]$, and $ \psi(\pi') = \psi(\pi) + \mathbb{E}_{\tau \sim \pi'}[\sum_{t=0}^\infty \gamma^t A_{\pi}(s_t, a_t)]$. In practice, the dependency on trajectories induced by $\pi'$ makes the above equation hard to optimize. To address this, the authors introduce a surrogate local approximation $L_\pi(\pi')$ to $\psi(\pi')$, using the state distribution over the current policy $\pi$ rather than $\pi'$:
\begin{equation}
\begin{split}
    L_\pi(\pi') &= \psi(\pi) + \sum_s \rho_\pi(s) \sum_a \pi'(a \vert s) A_\pi(s,a) \\
    &= \psi(\pi) + \mathbb{E}_{\tau \sim \pi} \Bigg[\sum_{t=0}^\infty \gamma^t \Tilde{A}(s_t) \Bigg]. 
\label{eq2}
\end{split}
\end{equation}
With the above intuitions, they derive an upper bound on the absolute difference between the objectives:
\begin{equation}
\begin{split}
\vert \psi(\pi') &- L_\pi(\pi')\vert \leq \frac{4\alpha^2 \gamma k}{(1-\gamma)^2}
\quad \text{with}~k = \max\limits_{s,a}\vert A_\pi(s,a) \vert, \\
\alpha &= D^{\text{max}}_{KL}(\pi, \pi') = \max\limits_{s} D_{KL}(\pi(\cdot \vert s) \;\vert \vert\; \pi'(\cdot \vert s)).
\end{split}
\end{equation}
Finally, by employing the relationship between the total variation (TV) divergence and the Kullback–Leibler (KL) divergence $D_{TV}(p\vert\vert q)^2 \leq D_{KL}(p\vert\vert q)$ \citep{pollard2000asymptopia}, \citet{TRPO} prove the following lower bound on the policy improvement:  
\begin{equation}
\psi(\pi') \geq L_\pi(\pi') - CD^{\text{max}}_{KL}(\pi, \pi'),\;\text{with}~C=\frac{4k\gamma}{1-\gamma^2}.
\end{equation}

\noindent\textbf{Exploration in RL.} In addition to the \textit{vine} TRPO method discussed in the previous section, a range of works investigate the idea of changing the initial state distribution \citep{2018-TOG-deepMimic, Ecoffet_2021, Messikommer24icra}. However, these works focus on improving exploration towards achieving higher returns rather than enforcing specific desired behaviors. For example, \citet{Messikommer24icra} uses states from past experiences to guide the agent toward states with higher payoffs. Similarly, \citet{Ecoffet_2021} stores and revisits promising states to explore the environment more efficiently. In contrast, \ourmethod: (i) focuses on refining agent behavior by repeatedly training on states where it failed to adhere to specific preferences, which makes it more applicable in tasks where behavior consistency and safety are required; and (ii) provides a lower bound on the policy improvement for mixed restart state distributions. For this reason, we believe our method is more closely related to CMDP-related literature (over which we compare in Section \ref{sec:experiments}) that is discussed in the following section.

\subsection{Constrained MDP}
Constrained RL encourages a behavioral preference, or a safety specification such as the ones we consider in our work \citep{ray2019safetygym, stooke2020lagpid, julien2022behavioralspec}. To this end, the classical MDP extends to a constrained MDP (CMDP) considering an additional set of $\mathcal{C} := \{C_i\}_{i\in n}$ indicator cost functions and $\mathbf{l} \in \mathbb{R}^n$ hard-coded thresholds for the constraints \citep{altman1999cmdp}. The goal of constrained RL algorithms is to maximize the expected reward while limiting the accumulation of costs under the thresholds. To this end, policy optimization algorithms typically employ the Lagrangian to transform the problem into an unconstrained one that is easy to implement over existing algorithms \citep{nocedal2006book}. 

Consider the case of a single constraint characterized by a cost function $C: \mathcal{S}\times\mathcal{A} \to \{0, 1\}$---define the expected cost function $\psi_C(\pi) := \mathbb{E}_{\tau\sim\pi}\left[\sum_{t=0}^\infty \gamma^t C(s_t, a_t) \right],$ and a cost threshold $l$.
The Lagrangian applies a differentiable penalty $\mathcal{L}_{\mathcal{C}}(\lambda)  = -\lambda\left(\psi_C(\pi) - l\right)$ to the policy optimization objective, where $\lambda$ is the so-called \textit{Lagrangian multiplier}. These algorithms thus take an additional gradient descent step in $\lambda$: $\nabla_\lambda \mathcal{L}_{\mathcal{C}}(\lambda) = l - \psi_C(\pi)$. The multiplier is forced to be $\geq 0$ as it acts as a penalty when the constraint is not satisfied (\textit{i.e.}, $\lambda$ increases) while decreasing to $0$ and removing any penalty when the constraint holds. However, choosing arbitrarily small values for the threshold potentially causes a detrimental trade-off between the main task and cost objectives, ultimately leading to policies that fail to solve the problem for which they are trained. Moreover, the cost metric employed in the safety evaluation is purely empirical and does not provide any provable guarantees on the actual adherence to behavioral preferences. To address these issues, we leverage formal verification of neural networks.

\subsection{Formal Verification of Neural Networks}
\label{sec:verification}
\textit{FV} is relevant to our work since it allows us to formalize a behavioral preference and provide provable guarantees on the adherence to these preferences. A reachability-based FV for neural networks tool takes as input a tuple $\mathcal{T} = \langle \mathcal{F}, \mathcal{X}, \mathcal{Y}\rangle$, where $\mathcal{F}$ 
\begin{figure}[b]
    \vspace{-8pt}
    \centering
    \includegraphics[width=.7\linewidth]{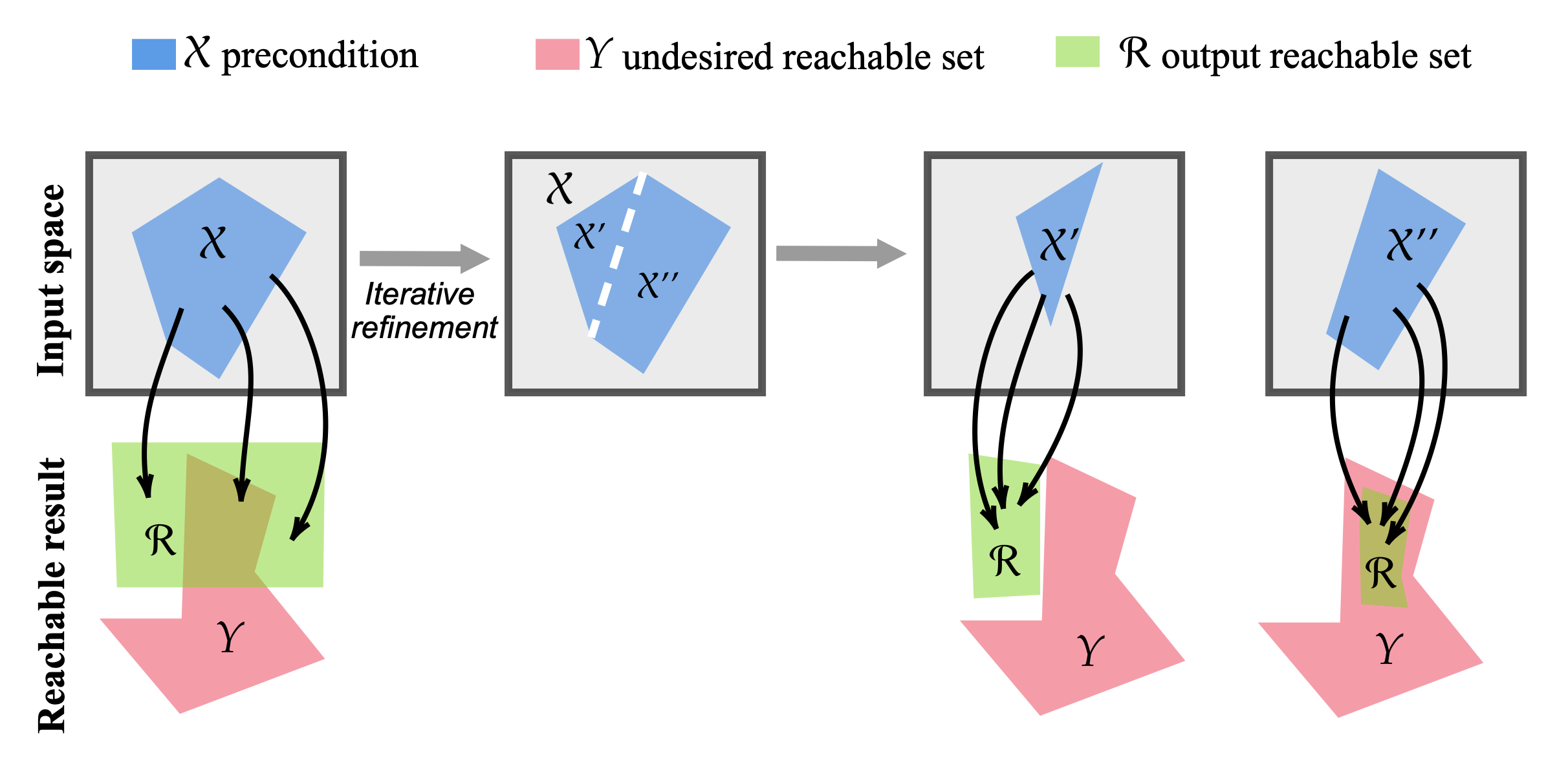}
    \vspace{-5pt}
    \caption{Overview of FV for neural networks.}
    \label{fig:overviewFV}
    \vspace{-8pt}
\end{figure}
is the trained policy (i.e., the neural network), and $\langle \mathcal{X}, \mathcal{Y}\rangle$ encodes a behavioral preference in terms of input-output relationships \citep{LiuSurvey}. 
Specifically, $\mathcal{X}$ is a precondition defined on the portion of the state space we are interested in, and $\mathcal{Y}$ models the postcondition specifying the desiderata. 
An FV tool propagates intervals $\mathcal{X}$ through $\mathcal{F}$ and performs a layer-by-layer reachability analysis to compute the output reachable set $\mathcal{R}(\mathcal{X}, \mathcal{F})$. The tool then checks if $\mathcal{R}(\mathcal{X}, \mathcal{F}) \subseteq \mathcal{Y}$, meaning that the agent satisfies the preference for all the states in $\mathcal{X}$.
Figure \ref{fig:overviewFV} shows a simplified overview of the verification process, which checks if (at least) one violation of the behavioral preference exists in $\mathcal{X}$. Due to over-approximation errors introduced by the propagations, FV tools iteratively split $\mathcal{X}$ into sub-domains $\mathcal{X}_i$ (the first two blocks in the figure) \citep{reluval}. When the output reachable set $\mathcal{R}(\mathcal{X}_i, \mathcal{F})$ is not included in $\mathcal{Y}$ (the second to last block), the iterative procedure ends---the behavioral preference is violated if at least one portion of the domain $\mathcal{X}$ falls within this scenario. As a natural extension of the FV problem, recent works \cite{CountingProVe,patching,eProve} propose to enumerate all the portions of $\mathcal{X}$ violating the desiderata, thus provably quantifying the rate at which agents satisfy the input-output relationships. In this work, we rely on the tool proposed by \cite{CountingProVe} to quantify the degree to which agents adhere to behavioral preferences in the explanatory robotic navigation task. 
\section{Policy Optimization via \ourmethod}
\label{sec:method_theory}
We introduce \ourmethod~to restart an agent from regions of the state space where it previously violated a behavioral preference. Our goal is to encourage a policy to exhibit behaviors aligned with the preference while improving performance and sample efficiency. To this end, \ourmethod~collects retrain areas---subsets of the state space $\overline{\mathcal{S}} \subseteq \mathcal{S}$ defined using an iterative procedure that merges parts of $\mathcal{S}$ where the agent violated the preference during training. We then introduce a \textit{mixed restart distribution}, combining the typical uniform restart distribution $\rho$ over the entire state space $\mathcal{S}$, with another uniform restart distribution $\overline{\rho}:\overline{\mathcal{S}} \to [0, 1]$ that considers retraining areas. Crucially, such a procedure is simple to implement and can potentially be applied to any RL algorithm. 

\begin{figure}[t]
    \centering
    \includegraphics[width=.6\linewidth]{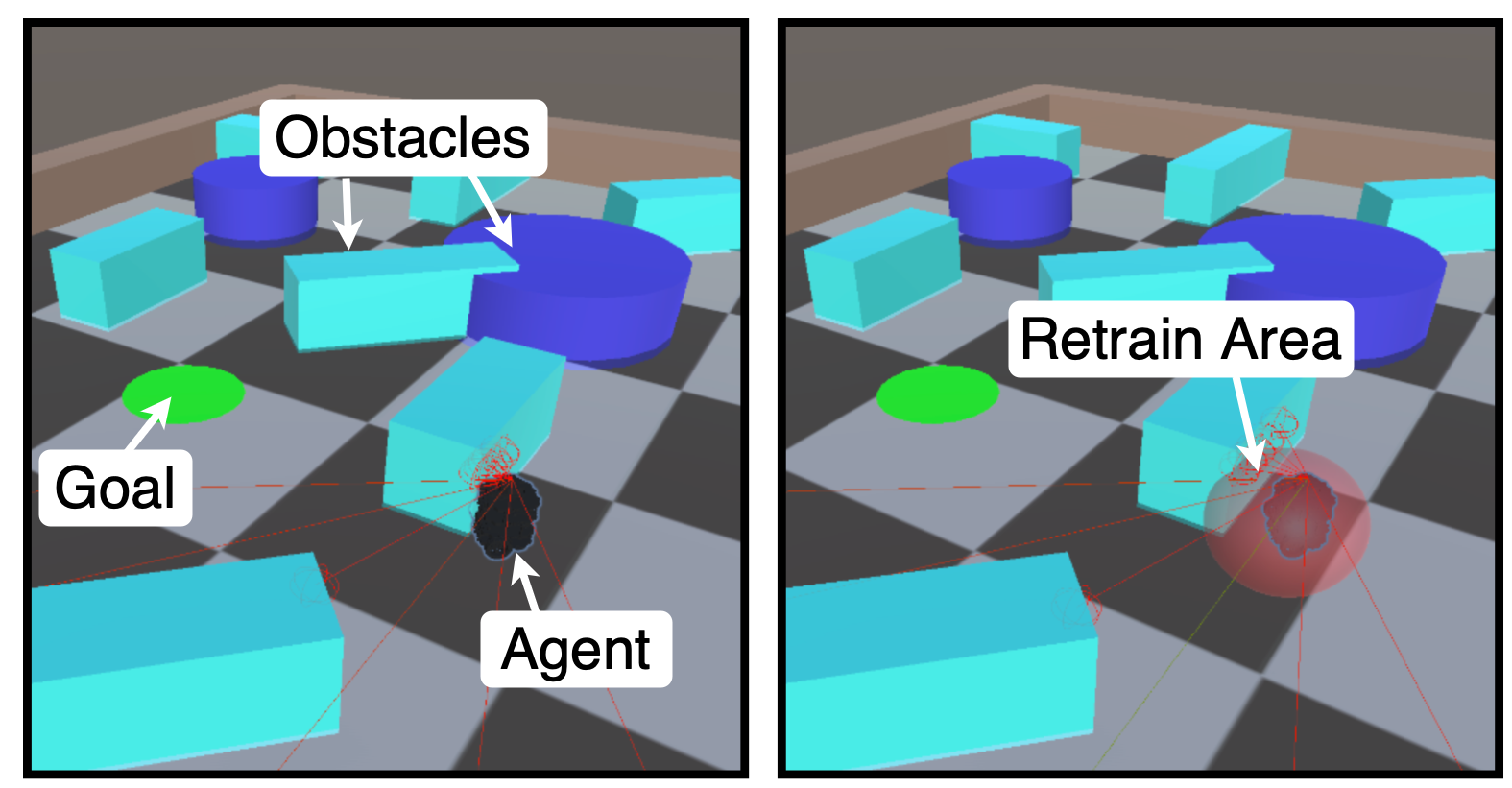}
    \caption{(left) The agent collides with an obstacle, receiving a positive cost. (right) A retrain area is created from that state.}
    \label{fig:overview_crop}
\end{figure}

Algorithm \ref{alg:eps_retrain} presents the pseudocode for \ourmethod. 
\begin{algorithm}[b]
\small
\caption{Template for \ourmethod~ methods}\label{alg:eps_retrain}
\begin{algorithmic}[1]
\Require \textit{bubble} size $\omega$ for initial retrain area, \textit{similarity} value $\beta$ to merge similar areas, \textit{decay}, \textit{initial}, and \textit{minimum} values for the $\varepsilon$ scaling. 
\State $\varepsilon\_decay  \gets (min\_\varepsilon -1.0)/(decay\cdot(epochs    \cdot steps\_per\_epoch))$
\State $\overline{\mathcal{S}} \gets \emptyset$;~~$s_0 \gets \rho(\mathcal{S})$ \Comment{Initialize areas buffer and environment}
\For{each episode}
\While{episode is not done}
    \State Execute the training loop of the RL algorithm for each step $t$.
    \If{$s_{t}$ is unsafe (i.e., cost $> 0$)}
         \State $r \gets \texttt{generate\_retrain\_area}(s_{t-1}, \omega)$
         \If{$\exists\; r'\ \in\overline{\mathcal{S}}$ \text{that is} $\beta$-similar to $r$}
            \State $r \gets \texttt{area\_refinement}(r,r')$
         \EndIf
         \State $\overline{\mathcal{S}} \gets \overline{\mathcal{S}} \cup r$
    \EndIf    
\EndWhile
\If{$random(0,1) < \varepsilon \wedge \overline{\mathcal{S}} \neq \emptyset$}
    \State $s_0 \gets \overline{\rho}(\texttt{sample\_retrain\_area}(\overline{\mathcal{S}}))$ 
    \State $\varepsilon \gets  \max(\varepsilon\_decay\cdot(epoch)*steps\_per\_epoch + 1.0,\; min\_\varepsilon)$
\Else
    \State $s_0 \gets \rho(\mathcal{S})$
\EndIf
\EndFor
\end{algorithmic}
\end{algorithm}
We start by initializing the memory buffer of the retrain areas $\overline{\mathcal{S}}$ as an empty set, and we select the starting state $s_0$ using the initial uniform state distribution $\rho$ over the entire state space $\mathcal{S}$ (line 2). In the training loop, the iterative procedure for collecting and merging retrain areas begins upon each unsafe interaction which returns a positive cost signal to the agent. We use this indicator cost signal as in the safe RL literature to detect the interactions where the agent violates the desiderata \citep{garcia2015safety}. Specifically, we generate a retrain area calling the \texttt{generate\_retrain\_area} method, which requires the previous state $s_{t-1}$ and generate an $\omega$-bubble (line 3-7) \citep{CROP}. Broadly speaking, such an area is a portion of the space surrounding the state that led to a violation, created by encoding each feature of the agent's state as an interval with fixed size $\omega$ (i.e., a ``bubble'' around the state). Figure \ref{fig:overview_crop} shows an explanatory example of an area for navigation, where collision avoidance is our desiderata. In this example, by leveraging commonly available information (e.g., the sensors' precision), we encode a bubble of size $\omega$ around the state that led to the collision. The idea is that retraining an agent from these similar collision-prone situations improves performance. 

This bubble thus becomes a retrain area $r$, and our approach automatically checks for the existence of another similar retrain area $r'$ to merge with, using the similarity threshold $\beta$ provided as a parameter. If such an $r'$ exists, the \texttt{area\_refinement} method is called (line 9). If no similar retrain area is found, $r$ is stored in the buffer $\overline{\mathcal{S}}$ that is initially empty (line 11). The refinement offers two advantages: (i) it allows us to maintain a reasonable size for $\overline{\mathcal{S}}$, and (ii) it clusters similar behavioral violations within the same retrain area, guaranteeing a uniform sampling over different violations. The new or refined area is then inserted into $\overline{\mathcal{S}}$ and can be used to retrain the agent. We refer to the next section for a complete overview of the generation and refinement methods. If there is at least one retrain area in $\overline{\mathcal{S}}$, the new initial state of the environment is either randomly sampled from the entire state space $\mathcal{S}$ with probability $1-\varepsilon$, or randomly sampled from a retrain area with probability $\varepsilon$ (lines 14-15). \ourmethod\ employs a linear decay for the mixed restarting distribution, avoiding the problem of getting stuck in suboptimal restart distributions (lines 16-20). If the sample is from a retrain area, the environment resets to a configuration (similar to) where the agent previously violated the behavioral preference. 

\subsection{Generation and Refinement Processes}

This section introduces the generation and refinement methods through a practical example. We first show the retrain area generation in the robotic navigation context and then the refinement process in the \textit{HalfCheetah} locomotion task. 

\subsubsection{Retrain Area Generation.} 
\begin{figure}[t]
    \centering
    \begin{minipage}[b]{0.31\textwidth}
        \centering
        \includegraphics[width=\textwidth]{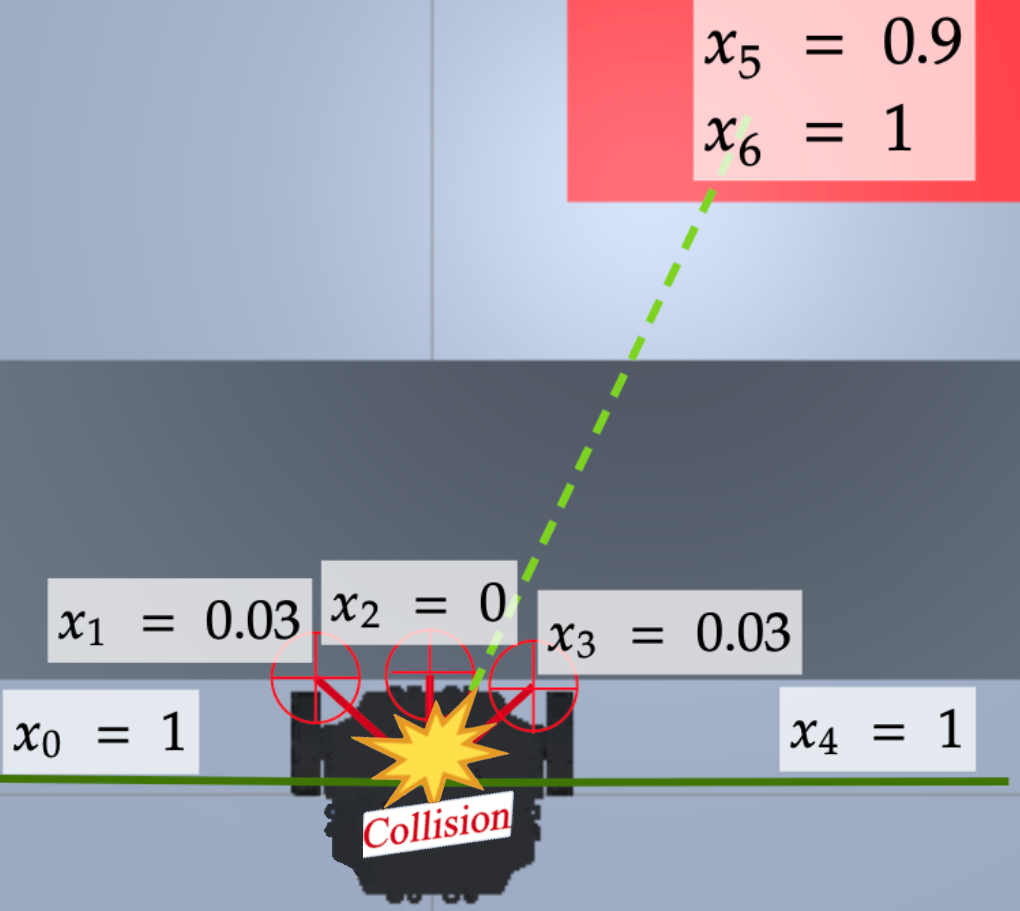}
        \label{fig:collision}
        \text{(a)}
    \end{minipage}
    \hfill
    \begin{minipage}[b]{0.31\textwidth}
        \centering
        \includegraphics[width=\textwidth]{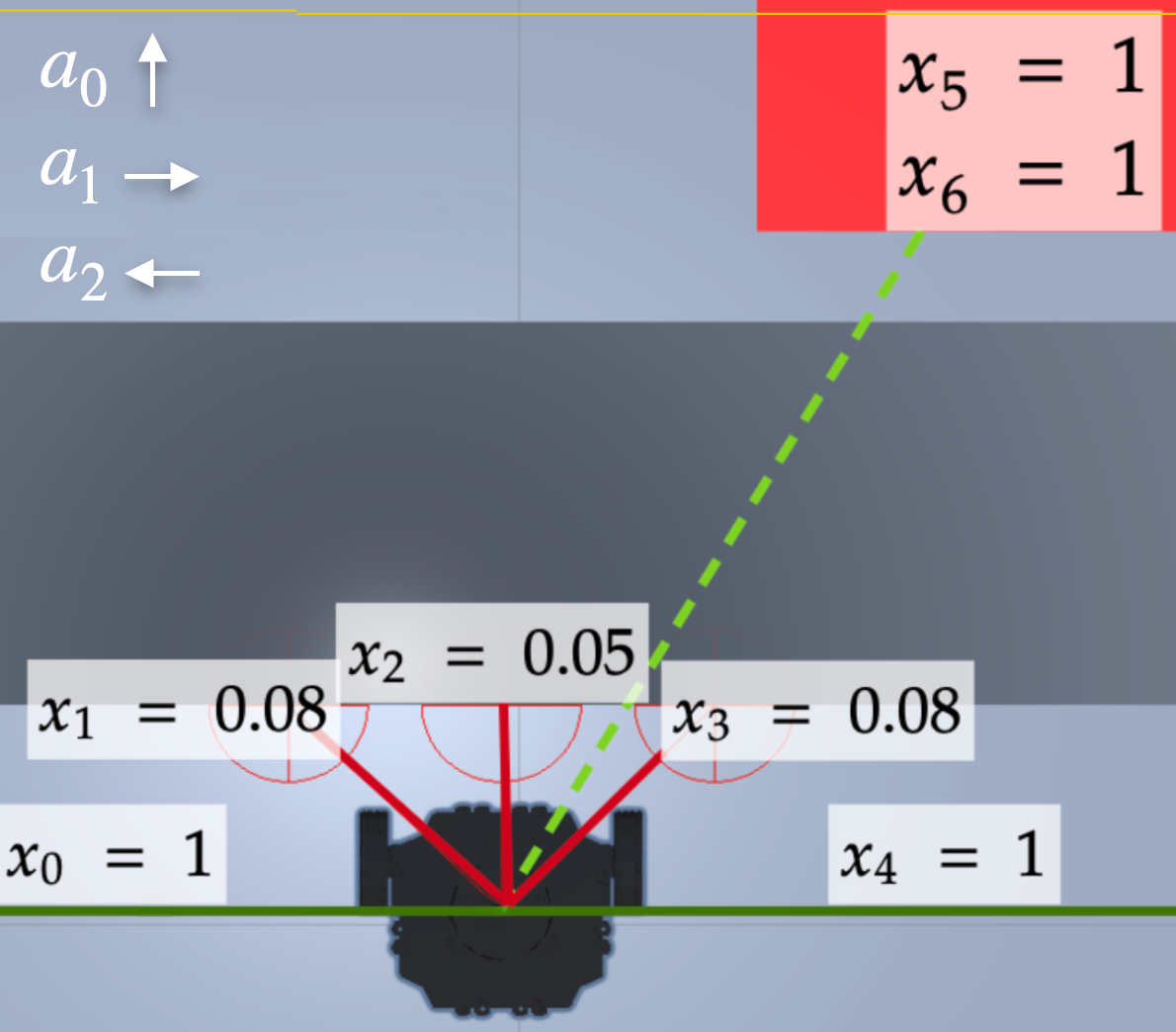}
        \label{fig:prev_state}
        \text{(b)}
    \end{minipage}
    \hfill
    \begin{minipage}[b]{0.31\textwidth}
        \centering
        \includegraphics[width=\textwidth]{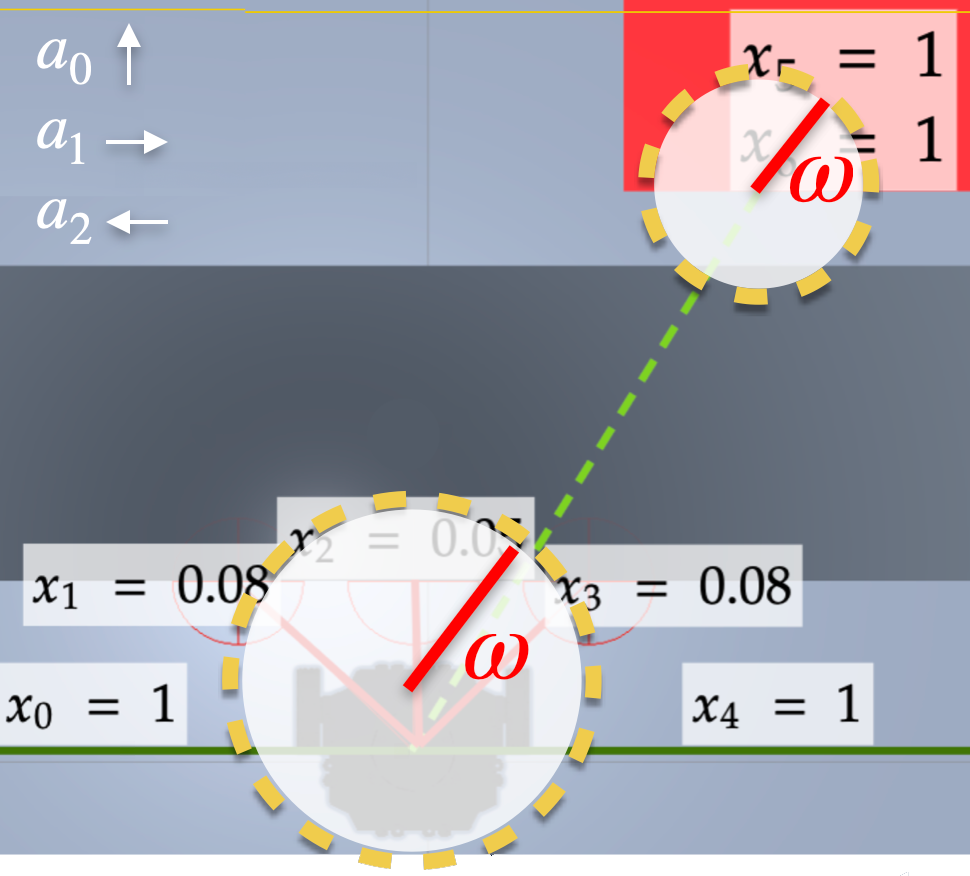}
        \label{fig:retrain_area}
        \text{(c)}
    \end{minipage}
    \caption{Retrain area generation. (a) Collision with an obstacle. (b) A previous unsafe state led to the collision. (c) $\omega$-bubble size to initialize the retrain area. Note that the $\omega$-bubble is the same for all the input features and is depicted in different sizes just for clarity representation purposes.}
    \label{fig:retrain_area_generation}
\end{figure}

Suppose an agent in a navigation scenario receives a positive cost signal from the environment. This indicates a collision with an obstacle, as depicted in Figure \ref{fig:retrain_area_generation}(a). The generation procedure selects the previous state $s_{t-1}$ that led to the collision as reported in Figure \ref{fig:retrain_area_generation}(b) and considers an $\omega$-bubble around this state to generate a retrain area as in Figure \ref{fig:retrain_area_generation}(c).
Taking $\omega=0.05$, i.e., a small value that encodes the surroundings of an unsafe situation, we obtain one interval for each input feature as:
\begin{align*}
    X: \{&x_0=[0.95,1], x_1=[0.03, 0.08], x_2=(0, 0.05],\\ &x_3=[0.03, 0.08],  x_4,x_5,x_6=[0.95,1]\}.
\end{align*}
We assume the states sampled from $X$ are undesirable---potentially risky. Therefore, when the initial state $s_0$ is sampled from this region $X$ using \ourmethod, the agent can improve the policy by learning how to better satisfy the desired safe behavior.

\subsubsection{Refinement Procedure} Once a retrain area has been created, \ourmethod\;checks whether it is possible to perform a refinement with an existing retrain area. To this end, we check if the distance between each corresponding interval in two different selected retrain areas, $X$ and $X'$, is less than or equal to $\beta$---a similarity threshold parameter. Formally, let \( X = \{x_0, x_1, \dots, x_n\} \) and \( X' = \{x_0', x_1', \dots, x_n'\} \), where each \( x_i \) and \( x_i' \) are intervals that encode possible value for each feature $x_i$ ($\forall i \in \{0, \dots, n\})$. We use Moore's interval algebra \citep{moore} and define the \textit{distance} between two intervals as \( [\underline{x_i}, \overline{x_i}] \) and \( [\underline{x'_i}, \overline{x'_i}]\) as:
\[ d([\underline{x_i}, \overline{x_i}], [\underline{x'_i}, \overline{x'_i}]) = \max(|\underline{x_i} - \underline{x'_i}|, |\overline{x_i} - \overline{x'_i}|). \]

Hence, two sets of intervals \( X \) and \( X' \), i.e., two retrain areas, are similar if and only if:\footnote{Without loss of generality, we assume the l2-norm in the features space is a meaningful distance metric. Using different metrics in different scenarios (e.g., robotic manipulation) does not impact our refinement procedure.}
\[ \forall i \in \{0, \dots, n\}, \quad d(x_i, x_i') \leq \beta. \]

\begin{figure}[h!]
    \centering
    \begin{minipage}[b]{0.3\textwidth}
        \includegraphics[width=0.8\textwidth]{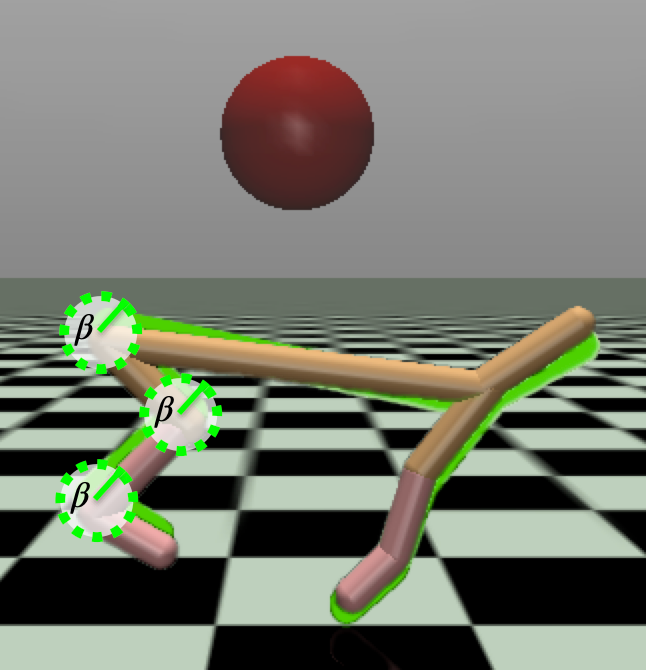}
    \end{minipage}
    \begin{minipage}[b]{0.3\textwidth}
        \includegraphics[width=0.8\textwidth]{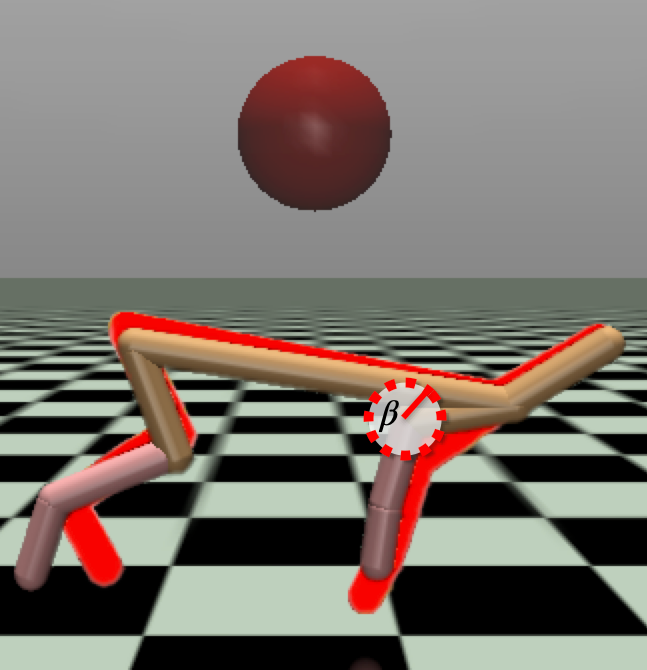}
    \end{minipage}
    \caption{Left: Explanatory similar unsafe states for a subset of the input features---the two states are within distance $\beta$. Right: Explanatory different unsafe situations---at least a couple of input features have a distance greater than $\beta$.}
    \label{fig:similarity}
\end{figure}

Figure \ref{fig:similarity} shows an example in a locomotion task, depicting a similar and not similar unsafe situation. This task has a desired velocity threshold for the Cheetah along the $x$ axis, and the red ball in the center of the image indicates a violation of such a threshold. For clarity, each figure uses two unsafe states (one with the original color and the other with a fixed red or green color) instead of intervals. If two sets of intervals \( X \) and \( X' \) are similar (left figure), \ourmethod\ combines these areas into a new area with intervals \( X'' \). For each pair of corresponding intervals \( x_i \in X \) and \( x'_i \in X' \) for all \( i \in \{0, \dots, n\} \), the new interval $x'' \in X''$ is computed as:

\[ x_i'' = [\min(\underline{x_i}, \underline{x'_i}), \max(\overline{x_i}, \overline{x'_i})], \]

meaning we take the minimum of the lower bounds and the maximum of the upper bounds of the corresponding intervals. Otherwise (right image), we keep the two areas separate.

\section{Policy Improvement}

In this section, we derive a bound on the monotonic policy improvement for \ourmethod's mixture of uniform restart distributions. We show that the original bound on monotonic policy improvement presented by \citet{TRPO} still holds and can be tighter. Notably, \textit{this result motivates both the design of our method as well as its superior performance} (Section \ref{sec:experiments}).
For the sake of clarity, we first recall the definition of \textit{$\alpha$-coupled policies} and the related lemma introduced to derive the bound in case of a single uniform restart distribution over $\mathcal{S}$. First, Definition \ref{def:coupling} couples two policies that behave in the same way (i.e., given a state, they pick the same action) with probability $\geq 1-\alpha$, and Lemma \ref{lem:bound} bounds the gap between policy advantages satisfying such policies.
\begin{definition}[$\alpha$-coupled policies \citet{TRPO}]\label{def:coupling}
    We say that $\pi$ and $\pi'$ are two $\alpha$-coupled policies if $\forall s \in\mathcal{S}$, we can define a joint distribution $(a, a') \vert s$ such that $P(a \neq a'\vert s) \leq \alpha$.
\end{definition}
\begin{lemma}[\citet{TRPO}] \label{lemma1}
    Given two $\alpha$-coupled policies, $\pi$ and $\pi'$, we have that: $\left\vert \mathbb{E}_{s_t\sim\pi'}[\Tilde{A}(s_t)] - \mathbb{E}_{s_t\sim\pi}[\Tilde{A}(s_t)]\right\vert \leq 4\alpha(1-(1-\alpha)^t)\max_{s,a}\vert A_\pi(s,a)\vert$.
    It follows that: $\vert \psi(\pi') - L_\pi(\pi')\vert$ $\leq \frac{4\alpha^2 \gamma k}{(1-\gamma)^2}$ with $k= \max\limits_{s,a}\vert A_\pi(s,a) \vert$.
\label{lem:bound}
\end{lemma}

We extend Lemma \ref{lem:bound} to the mixture of restarting distributions used by \ourmethod~(right side of Equation \ref{eq:mixture_return}, where $s_0\sim\overline{\rho}$). To this end, we define the expected discounted return of a new policy $\pi'$ over the current $\pi$ under the $\varepsilon$ mixture of restart policies as:
\begin{equation}
\begin{split}
     \overline{\psi}(\pi') =~&(1-\varepsilon)\Bigg[\psi(\pi) + \mathbb{E}_{ \substack{ \tau \sim \pi' \\ s_0 \sim \rho}} \Big[\sum_{t=0}^\infty \gamma^t \Tilde{A}(s_t) \Big]\Bigg] \\
     &+(\varepsilon)\Bigg[\psi(\pi) + \mathbb{E}_{ \substack{ \tau \sim \pi' \\ s_0 \sim \overline{\rho}}} \Big[\sum_{t=0}^\infty \gamma^t \Tilde{A}(s_t) \Big]\Bigg].
\label{eq:mixture_return}
\end{split}
\end{equation}

Hence, for the surrogate loss (see Equation \ref{eq2}), we compute the following local approximation:

\begin{equation} \label{eq2_new}
\begin{split}
\overline{L}_\pi(\pi') =~&(1-\varepsilon)\Bigg[\psi(\pi) + \mathbb{E}_{\substack{ \tau \sim \pi \\ s_0 \sim \rho }} \Big[\sum_{t=0}^\infty \gamma^t \Tilde{A}(s_t) \Big]\Bigg] \\
&+ (\varepsilon)\Bigg[\psi(\pi) + \mathbb{E}_{\substack{ \tau \sim \pi \\ s_0 \sim \overline{\rho}}} \Big[\sum_{t=0}^\infty \gamma^t \Tilde{A}(s_t) \Big]\Bigg].
\end{split}
\end{equation}

We then extend Lemma \ref{lem:bound} to provide an upper bound on the distance between $\overline{\psi}(\pi')$ and  $\overline{L}_\pi(\pi')$ under \ourmethod. 

\begin{lemma} \label{ourlemma}
    Considering a mixed restart distribution over $\rho$ and $\overline{\rho}$ using \ourmethod, it holds that
    $$\vert \overline{\psi}(\pi') - \overline{L}_\pi(\pi')\vert \leq \frac{4\alpha^2 \gamma}{(1-\gamma)^2}\Big[k(1-\varepsilon)+ k'\varepsilon\Big] \leq \frac{4\alpha^2 \gamma k}{(1-\gamma)^2},$$ with $k'= \max\limits_{\substack{
    s_0 \sim \overline{\rho}\\ \tau \sim \pi'}}\vert A_{\pi'}(s,a) \vert \leq k= \max\limits_{\substack{s_0 \sim \rho \\ \tau \sim \pi'}}\vert A_{\pi'}(s,a) \vert$, $\varepsilon \in [0,1]$.
\label{lem:epsbound}
\end{lemma}
\noindent\textit{Proof.} Full proof of Lemma \ref{lem:epsbound} is presented in Appendix \ref{app:proof}.\footnote{\href{https://arxiv.org/abs/2406.08315}{All the appendices are available at this link}.} $\quad\hfill \ensuremath{\Box}$ 
Finally, by exploiting the relationship between the total variation divergence and the KL divergence \citep{pollard2000asymptopia}, we derive the following corollary on the monotonic improvement guarantee under \ourmethod-based methods.
\begin{corollary}
    Let $\rho$ and $\overline{\rho}$ be two different restart distributions. Combining $\rho$ and $\overline{\rho}$ during the training as in \ourmethod, and by setting $\alpha=D_{KL}^{max}(\pi, \pi')$, the bound on the monotonic improvement for the policy update 
    $\overline{\psi}(\pi') \geq \overline{L}_\pi(\pi') - CD^{max}_{KL}(\pi, \pi')$ with $C=\frac{4k\gamma}{1-\gamma^2} $ 
    is still guaranteed.
\end{corollary}

The result naturally follows from Lemma \ref{ourlemma} and the original derivation of \citet{TRPO}.\footnote{We note the theoretically justified procedure motivating the policy improvement bound \citep{TRPO} does not hold for most practical deep RL policy optimization algorithms.}

\section{Limitations}

We identify the three following limitations in our work:
\begin{itemize}
    \item Our algorithm requires a simulator to train the agent and the possibility of resetting the system to specific states. We believe this requirement is reasonable in the RL literature.
    \item In \ourmethod~, we assume having access to an additional indicator cost signal from the environment. Such a signal is widely adopted in the safe RL literature, where system designers assume having access to a cost function that deems a state-action pair as safe or unsafe \citep{altman1999cmdp, ray2019safetygym, ji2023omnisafe}.
    \item We assume the area surrounding a collision state is also prone to violations of the desiderata. When such an assumption does not hold (e.g., in highly non-linear systems such as power grids), we use the exact feature values to determine a retrain area instead of intervals of size $\omega$.
\end{itemize}

\section{Experiments}
\label{sec:experiments}

\begin{figure}[h!]
    \centering
    \begin{minipage}[b]{0.15\textwidth}
        \centering
        \includegraphics[width=\textwidth]{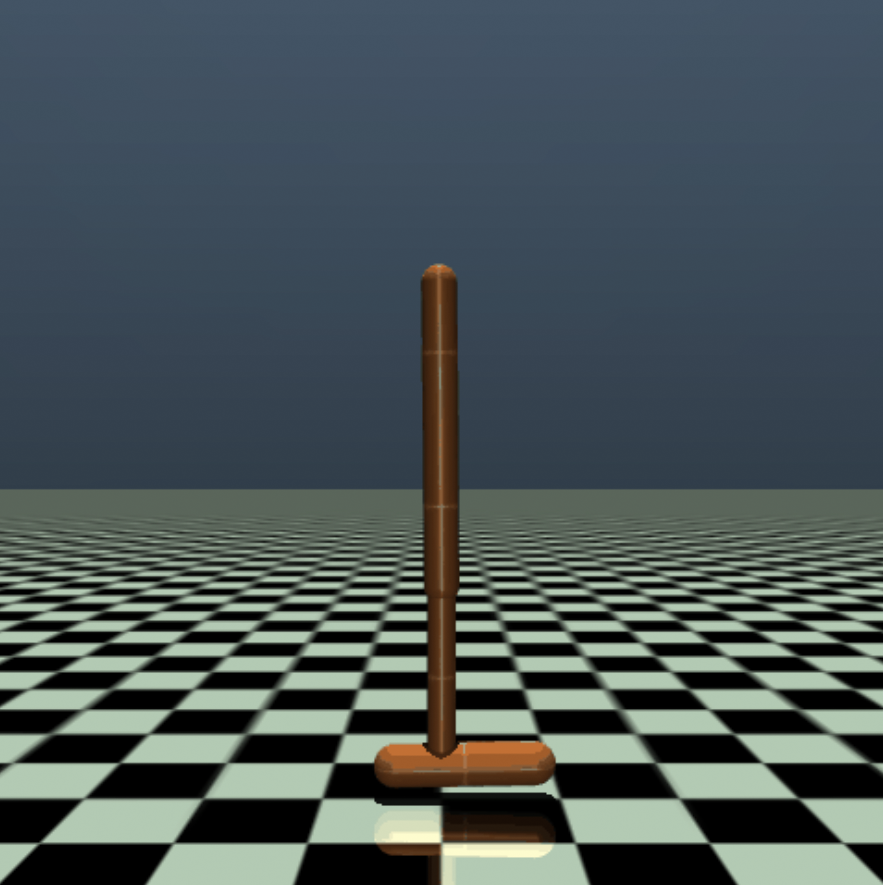}
        \label{fig:hopper}
        \text{(a) Hopper}
    \end{minipage}
    \begin{minipage}[b]{0.15\textwidth}
        \centering
        \includegraphics[width=\textwidth]{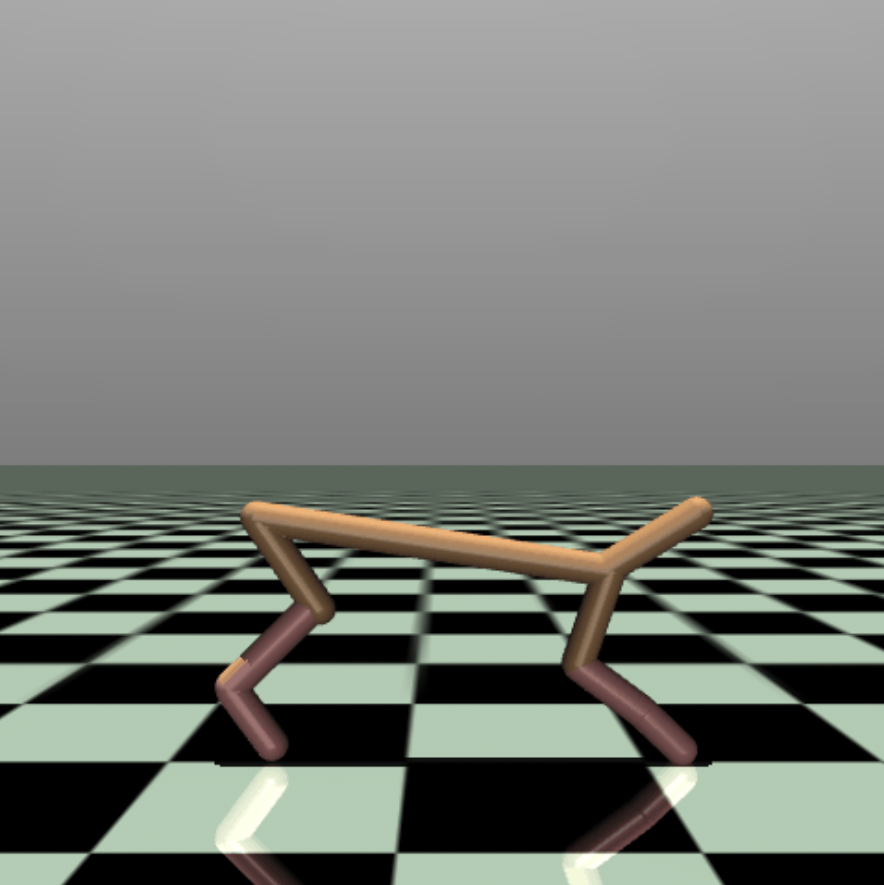}
        \label{fig:cheetah}
        \text{(b) HalfCheetah}
    \end{minipage}
    \begin{minipage}[b]{0.4\textwidth}
        \centering
        \includegraphics[width=\textwidth]{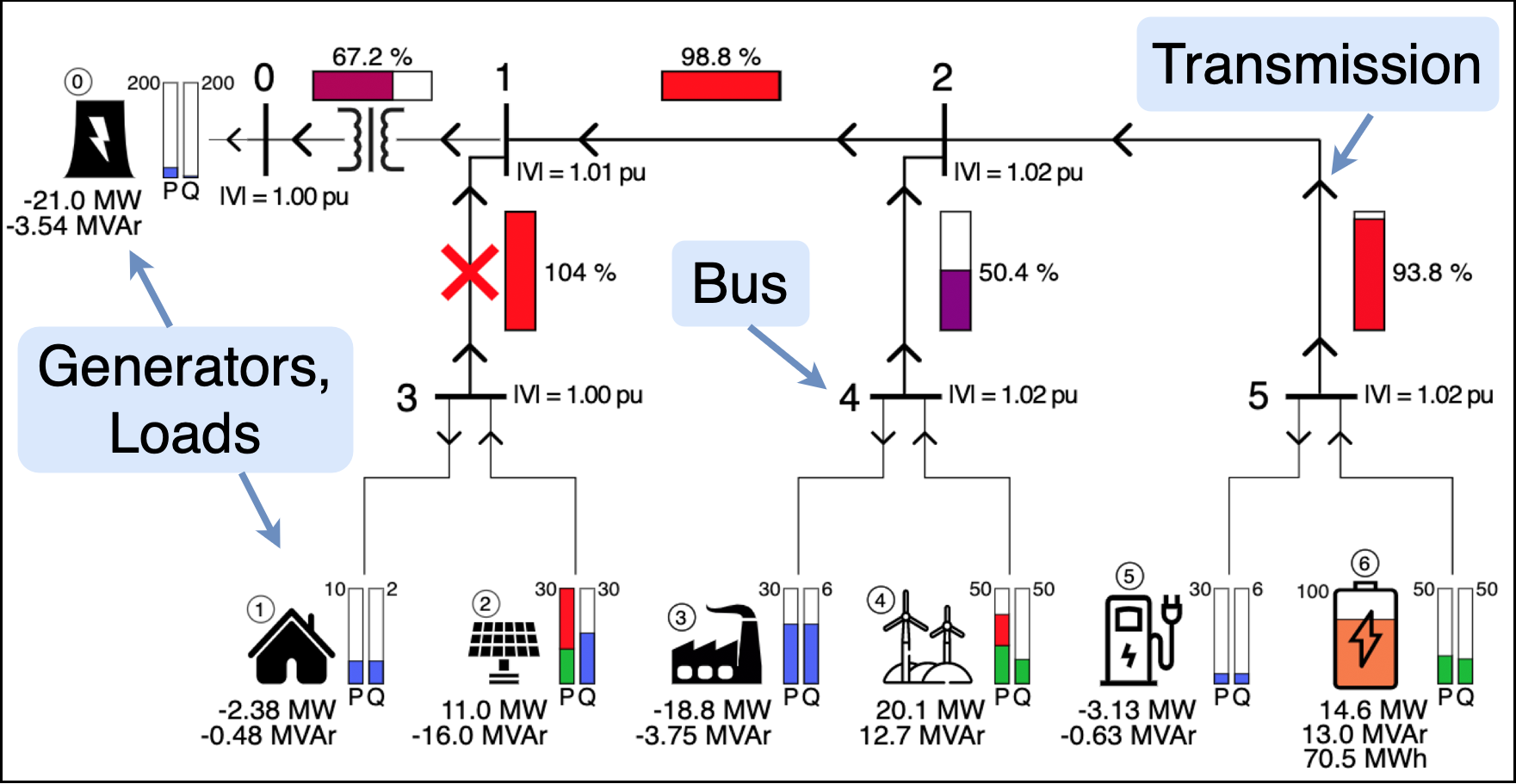}
        \label{fig:anm}
        \text{(c) Active Network Management}
    \end{minipage}
    \begin{minipage}[b]{0.24\textwidth}
        \centering
        \includegraphics[width=\textwidth]{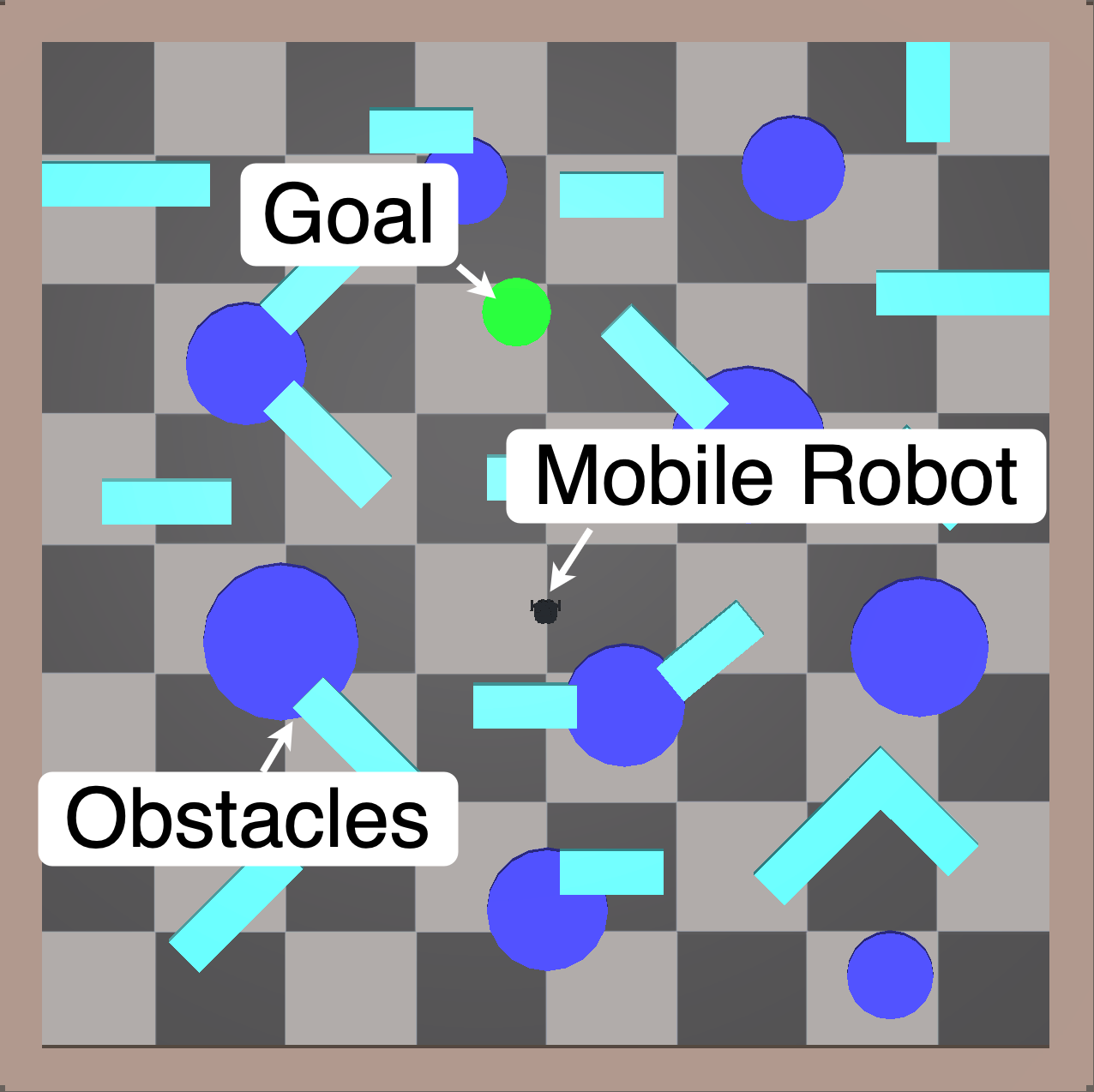}
        \label{fig:nav}
        \text{(d) Navigation}
    \end{minipage}
    \caption{Environments employed in our experiments.}
    \label{fig:envs}
\end{figure}

We \textit{present a comprehensive evaluation of \ourmethod~applied to TRPO} \citep{TRPO} that approximates the policy optimization theory of Section \ref{sec:method_theory}, \textit{and PPO} \citep{PPO} which relaxes the computational demands of TRPO.\footnote{We also tested \textit{vine} TRPO, which achieved comparable results with lower sample efficiency than TRPO. For this reason, our evaluation considers the TRPO algorithm.} We refer to these methods as $\varepsilon$-TRPO and $\varepsilon$-PPO. \textit{Additionally, we investigate the impact of the proposed approach on top and against safe RL baselines}, using the Lagrangian method with TRPO and PPO and modeling behavioral preferences as constraints. The resulting algorithms are named TRPOLagr, PPOLagr, $\varepsilon$-TRPOLagr, and $\varepsilon$-PPOLagr.
Our experiments address the following questions: 
\begin{itemize}
    \item Does \ourmethod~allow agents to better adhere to behavioral preferences while solving the task in both an unconstrained (where we penalize the reward upon violating the preference) and constrained formalization?
    \item How does \ourmethod~impact existing CMDP-based methods aimed at satisfying these preferences?
    \item How often do agents satisfy the behaviors provably and empirically?
\end{itemize}
To answer these questions, we begin our experiments using two known safety-oriented tasks, ``SafetyHopperVelocity-v1, and\\``SafetyHalfCheetahVelocity-v1", from the Safety-Gymnasium benchmark \citep{SafetyGymnasium}. To evaluate our method in a variety of setups and different behavioral desiderata, we also employ two practical scenarios based on an active network management task for a power system \citep{gymanm}, and navigation for a mobile robot \citep{pomcp_navigation,DUELMIX}. Figure \ref{fig:envs} shows these tasks. For simplicity, we will refer to them as \textit{Hopper}, \textit{HalfCheetah} (Cheetah), \textit{Active Network Management} (ANM), and \textit{Navigation}, respectively.
In the following, we briefly describe the tasks, referring to Appendix \ref{subsec:envs} for a more exhaustive description.

\begin{itemize}
    \item Hopper, Cheetah: The robots have to learn how to run forward by exerting torques on the joints and observing the body parts' angles and velocities (for a total of 12 and 18 input features). The actions control the torques applied to the (3 and 6) joints of the robot.
    \item ANM: The agent has to reschedule the power generation of different renewable and fossil generators, to satisfy the energy demand of three loads connected to the power grid. The agent observes the state of the power network through 18 features (i.e., active and reactive power injections, charge levels, and maximum productions) and controls power injections and curtailments using 6 continuous actions.
    \item Navigation: A mobile robot has to control its motor velocities to reach goals that randomly spawn in an obstacle-occluded environment without having a map. The agent observes the relative position of the goal and sparse lidar values sampled at a fixed angle (for a total of 22 features) and controls linear and angular velocity using 2 continuous actions.
\end{itemize}

\subsection{Implementation Details}
\label{sec:implementation_details}
Data collection is performed on Xeon E5-2650 CPU nodes with 64GB of RAM, using existing implementations for PPO, TRPO, and their Lagrangian version, based on the omnisafe library \citep{ji2023omnisafe}. Complete hyperparameters are in Appendix \ref{app:hyperparameters}. We report the average return, cost, and standard error as shaded regions over 50 independent runs per method. Figures~\ref{fig:base_results} and \ref{fig:lag_results} (the latter reported in the supplementary) show the average return in the first row and the average cost in the second row, where each column represents a different task. Notably, \textit{we are seeking agents that achieve a lower cost, which indicates they better adhere to the desired behavioral preferences while also solving the task}. Our claims on the performance improvement of \ourmethod~are supported by 1600 training runs, which significantly surpasses the typical 3-10 runs per method used in previous policy optimization works \citep{TRPO,PPO}. We note that due to employing a small penalty in the reward function to encourage specific behavioral preference, our results are not directly comparable to the published baselines \citep{TRPO, PPO}. For a fair comparison, we first collect the baseline with this new setting and then compare the performance with our approach. Considering the computational resources used for our extensive evaluation, Appendix \ref{app:env_impact} addresses the environmental impact of our experiments. 

\subsection{Empirical Evaluation}

\begin{figure*}[t]
    \centering
    \includegraphics[width=.95\linewidth]{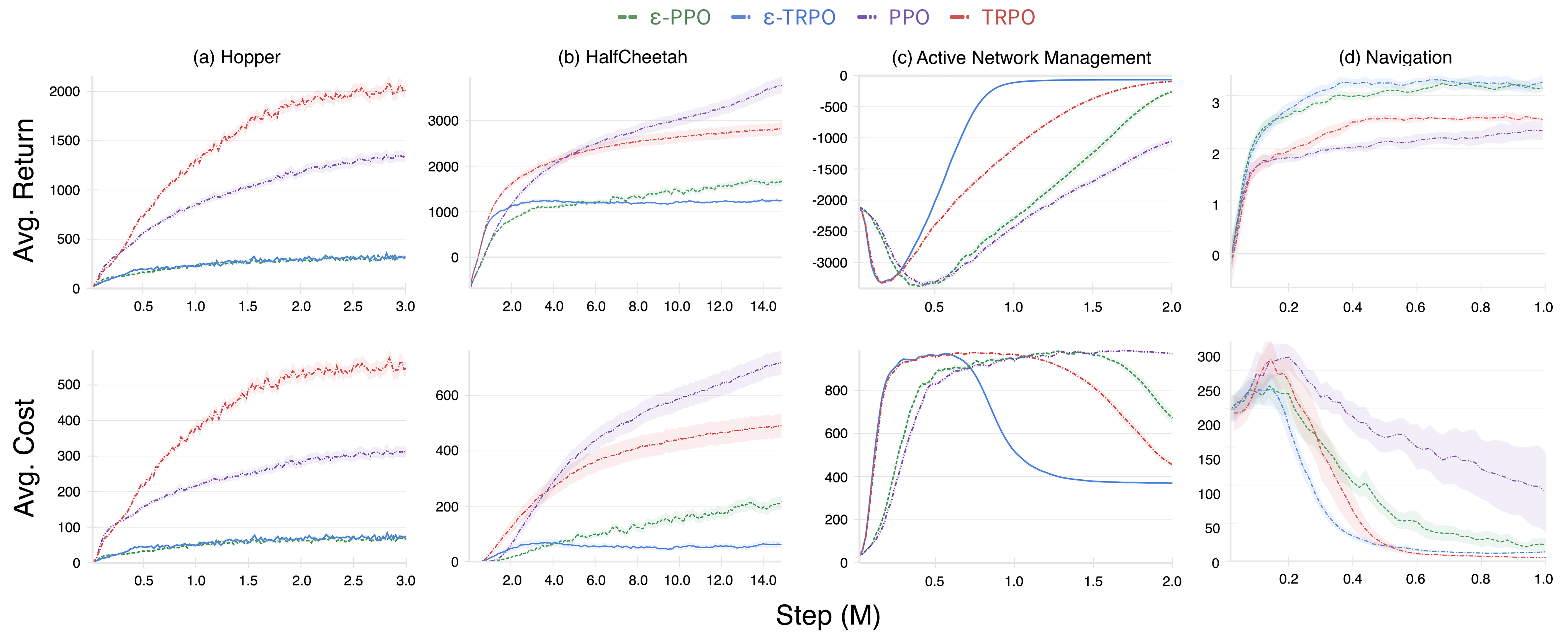}
    \vspace{-8pt}
     \caption{Comparison of $\varepsilon$-PPO, $\varepsilon$-TRPO, PPO and TRPO.} 
    \label{fig:base_results}
    \vspace{-5pt}
\end{figure*}

\textbf{Performance of $\varepsilon$-TRPO and $\varepsilon$-PPO.} Figure \ref{fig:base_results} shows that TRPO and PPO enhanced with our \ourmethod~improve sample efficiency and allow agents to better adhere to the behavioral preferences.

In Hopper and HalfCheetah, TRPO and PPO achieve substantially higher returns than their \ourmethod\;version; a result that could be easily misunderstood. In fact, this is related to the nature of the task, where the reward is directly proportional to the agents' velocities. 
For this reason, an agent that violates the behavioral preference \textit{``limit velocity under a threshold''}, achieves higher returns. This is clearly shown in the first two columns of Figure \ref{fig:base_results}, where $\varepsilon$-TRPO and $\varepsilon$-PPO resulted in notably lower cost compared to the baselines, indicating they lead the agents towards adhering to the velocity limit significantly more often than TRPO and PPO. Similar results are achieved in the ANM task where $\varepsilon$-TRPO and $\varepsilon$-PPO are notably safer and more sample efficient (see Pareto frontiers for convergence results in Figure \ref{fig:base_results_pareto} in the supplementary) than the baseline counterparts. 
The benefits of \ourmethod~are also confirmed in the navigation task, where violating the safety desiderata \textit{``avoid collisions''} leads to more collisions. Ultimately, achieving a higher cost (i.e., more collisions) hinders the navigation performance of the agent and leads to lower returns. Specifically, TRPO and $\varepsilon$-TRPO converge to the same average cost. However, the higher sample efficiency of the latter through the training, in terms of learning collision avoidance behaviors more quickly, allows $\varepsilon$-TRPO to learn better navigation behaviors, outperforming TRPO in terms of average return. Moreover, $\varepsilon$-PPO significantly outperforms PPO both in terms of average cost and return. 
\vspace{5pt}

\noindent\textbf{Performance of $\varepsilon$-TRPOLagr and $\varepsilon$-PPOLagr.} During training for the Lagrangian algorithms, both $\varepsilon$-TRPOLagr and $\varepsilon$-PPOLagr drastically reduce the amount of constraint violations in the Hopper and HalfCheetah velocity environments (complete learning curves are reported in Appendix \ref{suppl:results}). We specify the fraction of training steps where agents violate their constraint in Table \ref{tab:violation_perc}. 

\begin{table}[h!]
\centering
\begin{tabular}{ccccc}
\toprule
    & \textbf{Hopper} & \textbf{Cheetah} & \textbf{ANM} & \textbf{Navigation} \\ \midrule
\textbf{PPOLagr}                & 0.57   & 0.33        & \textbf{0.44}                      & 0.46       \\
\textbf{$\varepsilon$-PPOLagr}  & \textbf{0.04}   & \textbf{0}           & 0.59                      & \textbf{0.44}       \\\midrule
\textbf{TRPOLagr}               & 0.51   & 0.17        & \textbf{0.58}                      & 0.64       \\ 
\textbf{$\varepsilon$-TRPOLagr} & \textbf{0.25}   & \textbf{0}           & 0.79                      & \textbf{0.56}       \\
\bottomrule
\end{tabular}
\vspace{3mm}
\caption{Average fraction of the training steps where agents violate the constraints (lower is better).}
\label{tab:violation_perc}
\vspace{-5mm}
\end{table}

\begin{figure}[h!]
    \centering
    \includegraphics[width=0.7\linewidth]{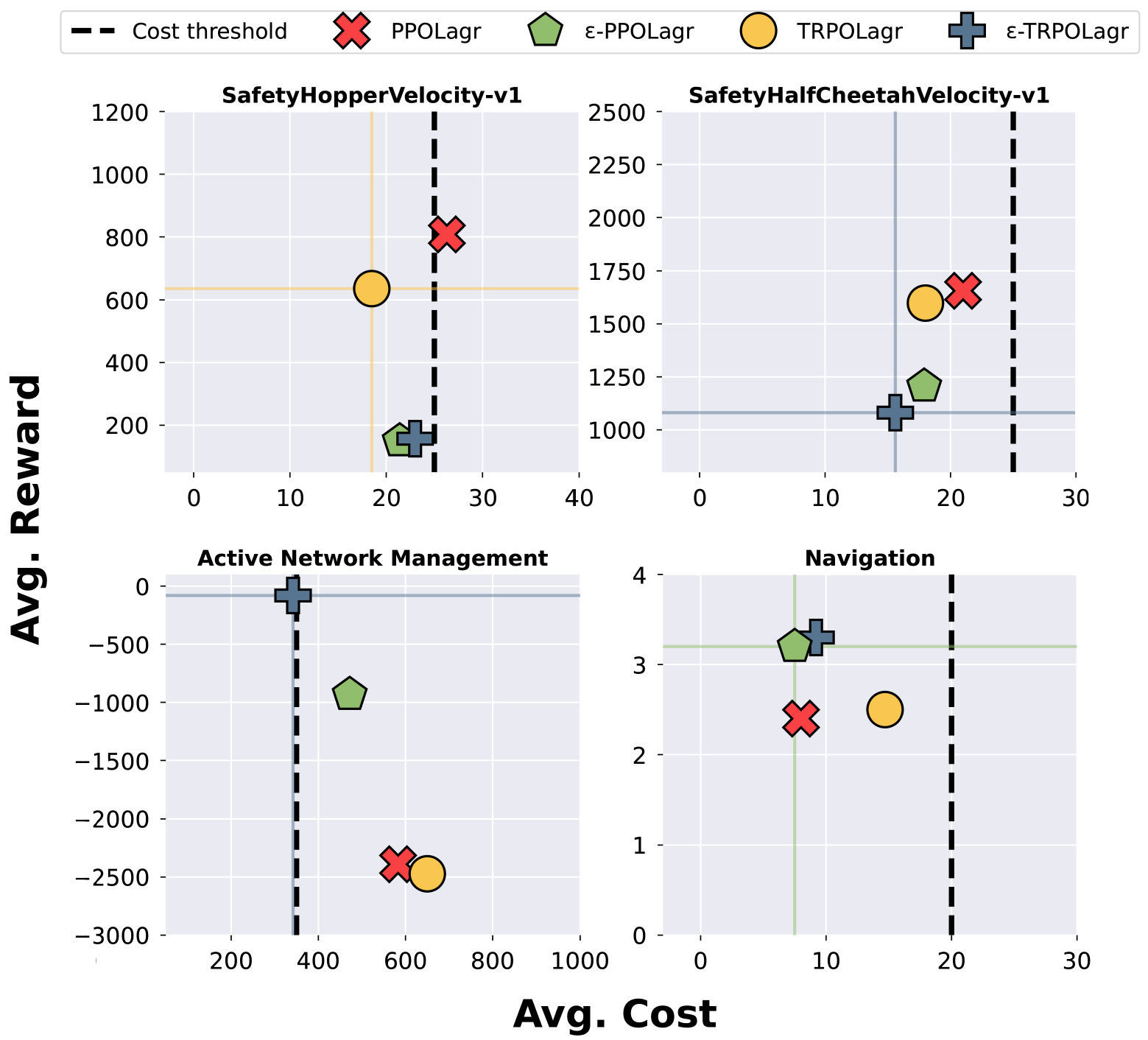}
    \vspace{-5pt}
    \caption{Pareto frontier of reward versus cost for $\varepsilon$-PPOLagr, $\varepsilon$-TRPOLagr, PPOLagr and TRPOLagr at convergence.}
    \vspace{-8pt}
    \label{fig:lag_results_pareto}
\end{figure}

However, at convergence, all the approaches satisfy the imposed thresholds. Figure \ref{fig:lag_results_pareto} shows the Pareto frontier reporting on the y-axis the average reward and on the x-axis the average cost at convergence. 
These results lead to some interesting considerations based on the setup of interest. 
In safety-critical contexts where it is crucial to satisfy constraints at training time, \ourmethod~showed significant empirical benefits. 

On the other hand, in non-critical contexts where performance at convergence is the main evaluation metric, the naive Lagrangian methods have superior return performance. Intuitively, this relates to the fact that Lagrangian methods often violate the constraints at training time, allowing agents to explore more and thus learn higher-performing behaviors.
In the more complex, realistic scenarios, our empirical analysis leads to different considerations. 
Specifically, in navigation, \ourmethod-based methods and the Lagrangian baselines achieve comparable results in terms of cost (i.e., constraint satisfaction). However, retraining agents in areas that are collision-prone allowed them to learn policies with better navigation skills and higher performance. 
In the ANM task, retraining an agent during grid instability increases the frequency of constraint violations compared to the baseline. Nonetheless, our approach helps agents learn to manage the grid effectively over time in contrast to Lagrangian baselines, which in the end, fail to solve the problem efficiently.

\subsection{Provably Verifying Navigation Behaviors} 
To further assess the benefit that \ourmethod~has over the behavioral preferences, we formally verify the policies trained for the navigation task. We consider this problem as an explanatory task for clarity since it allows us to easily visualize the retrain areas generated for \textit{``collision avoidance''} on top of the environment.

\begin{figure}[h!]
    \centering
    \includegraphics[width=0.6\linewidth]{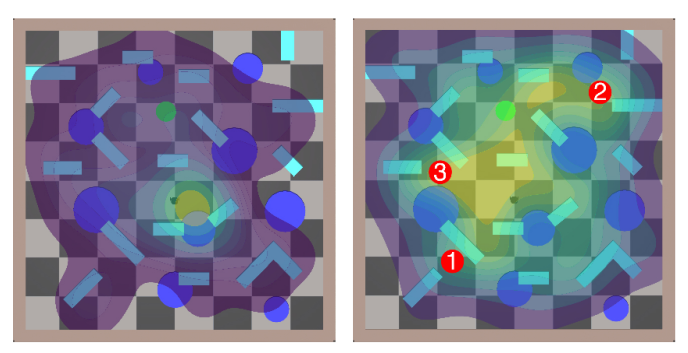}
    \caption{Density map of the retrain areas collected in the first and last training epochs; yellow indicates higher density.}
    \label{tab:resultsFV}
\end{figure}

Figure \ref{tab:resultsFV}, shows a kernel density estimation map of the retrain areas distribution at the beginning (left) and final stages (right) of the training for the $\varepsilon$-TRPO agent. Here we can notice how the agent successfully learns to navigate the environment over time since the retrain areas are more equally distributed through the entire scenario.
Using a recent verification tool \citep{CountingProVe}, we aim to quantify the probability that a navigation policy violates the collision avoidance preference. 
To this end, we consider three representative retrain areas collected while training the different \ourmethod-based algorithms (depicted as red dots in Figure \ref{tab:resultsFV}). For each area, we encode the input-output relationship required by the tool (see Section \ref{sec:verification}), considering the retrain area as the precondition, and the minimum linear and angular velocities that would cause a collision as the postcondition. Broadly speaking, the FV tool checks where the trained policies do not exceed such minimum velocities (i.e., they do not collide), and returns the portion of each retrain area for which the given policy violates the postcondition (i.e., the probability of colliding in that area). 
Table \ref{tab:fv} reports the probability that policies at convergence collide in the chosen retrain areas, averaged over all the runs. This additional FV-based analysis shows that \ourmethod~algorithms better adhere to the behavioral preference, further confirming our intuitions and the merits of our approach.  

\begin{table}[h!]
\centering
\begin{tabular}{llll}
\toprule
                       & \multicolumn{3}{l}{\textbf{Retrain areas (1, 2, 3)}} \\ \midrule
$\varepsilon$-PPO      & \textbf{0.007\%}        & \textbf{0.011\%}      & \textbf{0.22\%}      \\
PPO                    & 0.012\%        & 0.017\%      & 0.59\%      \\ \midrule
$\varepsilon$-TRPO     & \textbf{0.014\%}        & \textbf{0.67\%}       & 1\%         \\
TRPO                   & 0.015\%        & 0.69\%       & \textbf{0.8\%}       \\ \midrule
$\varepsilon$-PPOLagr  & \textbf{0.006\%}        & \textbf{0\%}          & \textbf{0.012\%}     \\
PPOLagr                & 0.013\%        & 0.05\%       & 0.1\%       \\ \midrule
$\varepsilon$-TRPOLagr & 0.0004\%       & 0\textbf{.007\%}      & \textbf{0.46\%}      \\
TRPOLagr               & \textbf{0.00005\%}      & 0.012\%      & 0.58\%      \\ \bottomrule
\end{tabular}
\vspace{3mm}
\caption{Average behavioral violations percentage for policies trained with TRPO, PPO, PPOLagr, TRPOLagr, and their \ourmethod\;version (ours).}
\label{tab:fv}
\end{table}

\subsection{Real (embodied) experiments} 
\begin{figure}[h!]
    \centering
    \includegraphics[width=.6\linewidth]{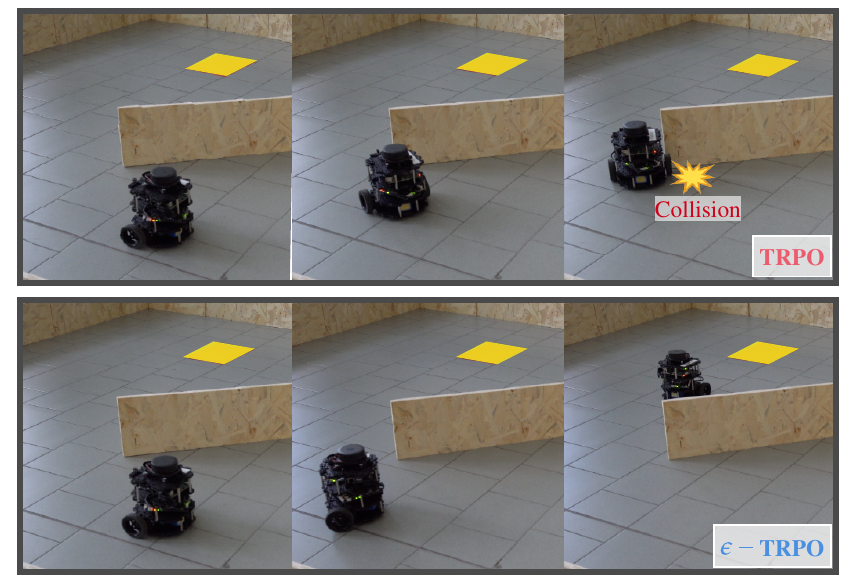}
    \caption{Real-world experiments comparing
    $\varepsilon$-TRPO and TRPO in corner-case scenarios. \href{https://youtu.be/rI21dUY43vI}{Video available here}.}
    \label{fig:real_exp}
\end{figure}

To conclude our comprehensive evaluation, we perform an additional evaluation in realistic (embodied) unsafe mapless navigation settings. Due to the similar performance at
convergence for TRPO and $\varepsilon$-TRPO in our simulated evaluations, we choose these two approaches for comparison. Specifically, we compare $\varepsilon$-TRPO and TRPO in scenarios where the agent has either all or only partially occluded LiDAR information.
We hypothesize that if the agent is not exposed to multiple unsafe situations during the training, i.e., without an \ourmethod\;strategy, it is less likely to select a longer but safer trajectory, and eventually, the agent will prefer a straight trajectory leading to a collision. 
To test this, using the Unity framework \cite{unity} we used to create the navigation task, we transfer the policies trained in simulation onto ROS-enabled platforms such as our Turtlebot3. We then test several corner-case situations, comparing the safer (from the formal verification results) trained agent at convergence for both $\varepsilon$-TRPO and TRPO. In our experiments, we observe our hypothesis to be correct (see Fig.~\ref{fig:real_exp}), showing the benefit of retraining the agent in specific regions of the state space that are deemed unsafe.

\section{Discussion}

This paper presented \ourmethod, a novel exploration strategy with monotonic improvement guarantees that optimizes policies while encouraging specific behavioral preferences. Our approach aims at retraining an RL agent from \textit{retrain areas} where it violated a desired behavioral preference at training time. 
Our empirical and formal evaluation over hundreds of seeds considering various tasks and behavioral preferences, demonstrated the effectiveness in terms of higher sample efficiency and superior performance of \ourmethod~when integrated with existing policy optimization methods. Real-world experiments confirmed the benefit of the proposed approach in realistic setups.

\section*{Acknowledgments}
This work was partly supported by mobility grants for non-EU destinations at the University of Verona's Doctoral School, the AI2050 program at Schmidt Sciences (Grant G-24-66236), and the MIT Climate Nucleus Fast Forward Faculty Fund Grant Program.



\bibliography{example}  

\begin{thebibliography}{38}
\providecommand{\natexlab}[1]{#1}
\providecommand{\url}[1]{\texttt{#1}}
\expandafter\ifx\csname urlstyle\endcsname\relax
  \providecommand{\doi}[1]{doi: #1}\else
  \providecommand{\doi}{doi: \begingroup \urlstyle{rm}\Url}\fi

\bibitem[Silver et~al.(2021)Silver, Singh, Precup, and Sutton]{silver2021reward}
D.~Silver, S.~Singh, D.~Precup, and R.~S. Sutton.
\newblock Reward is enough.
\newblock \emph{Artificial Intelligence}, 299:\penalty0 103535, 2021.
\newblock ISSN 0004-3702.

\bibitem[Amodei et~al.(2016)Amodei, Olah, Steinhardt, Christiano, Schulman, and Man{\'{e}}]{amodei2016rewards}
D.~Amodei, C.~Olah, J.~Steinhardt, P.~F. Christiano, J.~Schulman, and D.~Man{\'{e}}.
\newblock Concrete problems in {AI} safety.
\newblock \emph{arXiv preprint arXiv:1606.06565}, 2016.

\bibitem[Tai et~al.(2017)Tai, Paolo, and Liu]{drl_navigation1}
L.~Tai, G.~Paolo, and M.~Liu.
\newblock Virtual-to-real drl: Continuous control of mobile robots for mapless navigation.
\newblock In \emph{IROS}, 2017.

\bibitem[Zhelo et~al.(2018)Zhelo, Zhang, Tai, Liu, and Burgard]{drl_navigation2}
O.~Zhelo, J.~Zhang, L.~Tai, M.~Liu, and W.~Burgard.
\newblock Curiosity-driven exploration for mapless navigation with deep reinforcement learning.
\newblock \emph{arXiv preprint arXiv:1804.00456}, 2018.

\bibitem[Kakade and Langford(2002)]{kakade2002approximately}
S.~Kakade and J.~Langford.
\newblock Approximately optimal approximate reinforcement learning.
\newblock In \emph{Proceedings of the Nineteenth International Conference on Machine Learning}, pages 267--274, 2002.

\bibitem[Schulman et~al.(2015)Schulman, Levine, Abbeel, Jordan, and Moritz]{TRPO}
J.~Schulman, S.~Levine, P.~Abbeel, M.~Jordan, and P.~Moritz.
\newblock Trust region policy optimization.
\newblock In \emph{International conference on machine learning}, pages 1889--1897. PMLR, 2015.

\bibitem[Eysenbach et~al.(2018)Eysenbach, Gu, Ibarz, and Levine]{eysenbach2018reset}
B.~Eysenbach, S.~Gu, J.~Ibarz, and S.~Levine.
\newblock Leave no trace: Learning to reset for safe and autonomous reinforcement learning.
\newblock In \emph{International Conference on Learning Representations}, 2018.

\bibitem[Jiang et~al.(2023)Jiang, Kolter, and Raileanu]{jian2023generalization}
Y.~Jiang, J.~Z. Kolter, and R.~Raileanu.
\newblock On the importance of exploration for generalization in reinforcement learning.
\newblock In \emph{Thirty-seventh Conference on Neural Information Processing Systems}, 2023.
\newblock URL \url{https://openreview.net/forum?id=y5duN2j9s6}.

\bibitem[Lagoudakis and Parr(2003)]{lagoudakis2003reinforcement}
M.~G. Lagoudakis and R.~Parr.
\newblock Reinforcement learning as classification: Leveraging modern classifiers.
\newblock In \emph{Proceedings of the 20th International Conference on Machine Learning (ICML-03)}, pages 424--431, 2003.

\bibitem[Gabillon et~al.(2013)Gabillon, Ghavamzadeh, and Scherrer]{gabillon2013approximate}
V.~Gabillon, M.~Ghavamzadeh, and B.~Scherrer.
\newblock Approximate dynamic programming finally performs well in the game of tetris.
\newblock \emph{Advances in neural information processing systems}, 26, 2013.

\bibitem[Marchesini and Amato(2023)]{VFS}
E.~Marchesini and C.~Amato.
\newblock Improving deep policy gradients with value function search.
\newblock In \emph{The Eleventh International Conference on Learning Representations}, 2023.
\newblock URL \url{https://openreview.net/forum?id=6qZC7pfenQm}.

\bibitem[Schulman et~al.(2017)Schulman, Wolski, Dhariwal, Radford, and Klimov]{PPO}
J.~Schulman, F.~Wolski, P.~Dhariwal, A.~Radford, and O.~Klimov.
\newblock Proximal policy optimization algorithms.
\newblock \emph{arXiv preprint arXiv:1707.06347}, 2017.

\bibitem[Roy et~al.(2022)Roy, Girgis, Romoff, Bacon, and Pal]{julien2022behavioralspec}
J.~Roy, R.~Girgis, J.~Romoff, P.-L. Bacon, and C.~J. Pal.
\newblock Direct behavior specification via constrained reinforcement learning.
\newblock In \emph{ICML}, volume 162, pages 18828--18843, 2022.

\bibitem[Ji et~al.(2024)Ji, Zhang, Zhou, Pan, Huang, Sun, Geng, Zhong, Dai, and Yang]{SafetyGymnasium}
J.~Ji, B.~Zhang, J.~Zhou, X.~Pan, W.~Huang, R.~Sun, Y.~Geng, Y.~Zhong, J.~Dai, and Y.~Yang.
\newblock Safety gymnasium: A unified safe reinforcement learning benchmark.
\newblock \emph{Advances in Neural Information Processing Systems}, 36, 2024.

\bibitem[Stooke et~al.(2020)Stooke, Achiam, and Abbeel]{stooke2020lagpid}
A.~Stooke, J.~Achiam, and P.~Abbeel.
\newblock Responsive safety in reinforcement learning by pid lagrangian methods.
\newblock In \emph{ICML}, 2020.

\bibitem[Marchesini and Farinelli(2022)]{EPS}
E.~Marchesini and A.~Farinelli.
\newblock Enhancing deep reinforcement learning approaches for multi-robot navigation via single-robot evolutionary policy search.
\newblock In \emph{International Conference on Robotics and Automation (ICRA)}, pages 5525--5531, 2022.
\newblock \doi{10.1109/ICRA46639.2022.9812341}.

\bibitem[Aydeniz et~al.(2024)Aydeniz, Marchesini, Amato, and Tumer]{marl_navigation}
A.~A. Aydeniz, E.~Marchesini, C.~Amato, and K.~Tumer.
\newblock Entropy seeking constrained multiagent reinforcement learning.
\newblock In \emph{Proceedings of the 23rd International Conference on Autonomous Agents and Multiagent Systems}, page 2141–2143, 2024.
\newblock ISBN 9798400704864.

\bibitem[Pollard(2000)]{pollard2000asymptopia}
D.~Pollard.
\newblock Asymptopia: an exposition of statistical asymptotic theory.
\newblock In \emph{Asymptopia: an exposition of statistical asymp-totic theory}, 2000.

\bibitem[Peng et~al.(2018)Peng, Abbeel, Levine, and van~de Panne]{2018-TOG-deepMimic}
X.~B. Peng, P.~Abbeel, S.~Levine, and M.~van~de Panne.
\newblock Deepmimic: Example-guided deep reinforcement learning of physics-based character skills.
\newblock \emph{ACM Trans. Graph.}, 37\penalty0 (4):\penalty0 143:1--143:14, July 2018.

\bibitem[Ecoffet et~al.(2021)Ecoffet, Huizinga, Lehman, Stanley, and Clune]{Ecoffet_2021}
A.~Ecoffet, J.~Huizinga, J.~Lehman, K.~O. Stanley, and J.~Clune.
\newblock First return, then explore.
\newblock \emph{Nature}, 590:\penalty0 580–586, 2021.
\newblock ISSN 1476-4687.

\bibitem[Messikommer et~al.(2024)Messikommer, Song, and Scaramuzza]{Messikommer24icra}
N.~Messikommer, Y.~Song, and D.~Scaramuzza.
\newblock Contrastive initial state buffer for reinforcement learning.
\newblock \emph{2024 IEEE International Conference on Robotics and Automation (ICRA)}, 2024.

\bibitem[Ray et~al.(2019)Ray, Achiam, and Amodei]{ray2019safetygym}
A.~Ray, J.~Achiam, and D.~Amodei.
\newblock Benchmarking safe exploration in deep reinforcement learning.
\newblock In \emph{OpenAI Blog}, 2019.

\bibitem[Altman(1999)]{altman1999cmdp}
E.~Altman.
\newblock Constrained markov decision processes.
\newblock In \emph{CRC Press}, 1999.

\bibitem[Nocedal and Wright(2006)]{nocedal2006book}
J.~Nocedal and S.~Wright.
\newblock \emph{Numerical Optimization}.
\newblock Springer Series in Operations Research and Financial Engineering. Springer, 2 edition, 2006.
\newblock ISBN 9780387400655.

\bibitem[Liu et~al.(2021)Liu, Arnon, Lazarus, Strong, Barrett, Kochenderfer, et~al.]{LiuSurvey}
C.~Liu, T.~Arnon, C.~Lazarus, C.~Strong, C.~Barrett, M.~J. Kochenderfer, et~al.
\newblock Algorithms for verifying deep neural networks.
\newblock \emph{Foundations and Trends{\textregistered} in Optimization}, 4\penalty0 (3-4):\penalty0 244--404, 2021.

\bibitem[Wang et~al.(2018)Wang, Pei, Whitehouse, Yang, and Jana]{reluval}
S.~Wang, K.~Pei, J.~Whitehouse, J.~Yang, and S.~Jana.
\newblock Formal security analysis of neural networks using symbolic intervals.
\newblock In \emph{27th USENIX Security Symposium (USENIX Security 18)}, pages 1599--1614, 2018.

\bibitem[Marzari et~al.(2023)Marzari, Corsi, Cicalese, and Farinelli]{CountingProVe}
L.~Marzari, D.~Corsi, F.~Cicalese, and A.~Farinelli.
\newblock The \#{DNN}-{V}erification {P}roblem: {C}ounting {U}nsafe {I}nputs for {D}eep {N}eural {N}etworks.
\newblock In \emph{International Joint Conference on Artificial Intelligence (IJCAI)}, pages 217--224, 2023.

\bibitem[Yang et~al.(2022)Yang, Yamaguchi, Tran, Hoxha, Johnson, and Prokhorov]{patching}
X.~Yang, T.~Yamaguchi, H.-D. Tran, B.~Hoxha, T.~T. Johnson, and D.~Prokhorov.
\newblock Neural network repair with reachability analysis.
\newblock In \emph{Formal Modeling and Analysis of Timed Systems}, pages 221--236, 2022.

\bibitem[Marzari et~al.(2024)Marzari, Corsi, Marchesini, Alessandro, and Cicalese]{eProve}
L.~Marzari, D.~Corsi, E.~Marchesini, F.~Alessandro, and F.~Cicalese.
\newblock Enumerating safe regions in deep neural networks with provable probabilistic guarantees.
\newblock \emph{Proceedings of the AAAI Conference on Artificial Intelligence}, pages 21387--21394, 2024.

\bibitem[Garc{\i}a and Fern{\'a}ndez(2015)]{garcia2015safety}
J.~Garc{\i}a and F.~Fern{\'a}ndez.
\newblock A comprehensive survey on safe reinforcement learning.
\newblock In \emph{JMLR}, 2015.

\bibitem[Marzari et~al.(2023)Marzari, Marchesini, and Farinelli]{CROP}
L.~Marzari, E.~Marchesini, and A.~Farinelli.
\newblock Online safety property collection and refinement for safe deep reinforcement learning in mapless navigation.
\newblock In \emph{2023 IEEE International Conference on Robotics and Automation (ICRA)}, pages 7133--7139, 2023.

\bibitem[Moore(1962)]{moore}
R.~E. Moore.
\newblock Interval arithmetic and automatic error analysis in digital computing.
\newblock Technical report, Stanford Univ Calif Applied Mathematics And Statistics Labs, 1962.

\bibitem[Ji et~al.(2023)Ji, Zhou, Borong~Zhang, Xuehai~Pan, Huang, Geng, Liu, and Yang]{ji2023omnisafe}
J.~Ji, J.~Zhou, J.~D. Borong~Zhang, R.~S. Xuehai~Pan, W.~Huang, Y.~Geng, M.~Liu, and Y.~Yang.
\newblock Omnisafe: An infrastructure for accelerating safe reinforcement learning research.
\newblock \emph{arXiv preprint arXiv:2305.09304}, 2023.

\bibitem[Henry and Ernst(2021)]{gymanm}
R.~Henry and D.~Ernst.
\newblock Gym-anm: Reinforcement learning environments for active network management tasks in electricity distribution systems.
\newblock \emph{Energy and AI}, 5:\penalty0 100092, 2021.
\newblock ISSN 2666-5468.

\bibitem[Castellini et~al.(2020)Castellini, Marchesini, Mazzi, and Farinelli]{pomcp_navigation}
A.~Castellini, E.~Marchesini, G.~Mazzi, and A.~Farinelli.
\newblock Explaining the influence of prior knowledge on pomcp policies.
\newblock In \emph{Multi-Agent Systems and Agreement Technologies}, pages 261--276, 2020.
\newblock ISBN 978-3-030-66412-1.

\bibitem[Marchesini and Farinelli(2021)]{DUELMIX}
E.~Marchesini and A.~Farinelli.
\newblock Centralizing state-values in dueling networks for multi-robot reinforcement learning mapless navigation.
\newblock In \emph{IEEE/RSJ International Conference on Intelligent Robots and Systems (IROS)}, pages 4583--4588, 2021.
\newblock \doi{10.1109/IROS51168.2021.9636349}.

\bibitem[Juliani et~al.(2018)Juliani, Berges, Vckay, Gao, Henry, Mattar, and Lange]{unity}
A.~Juliani, V.~Berges, E.~Vckay, Y.~Gao, H.~Henry, M.~Mattar, and D.~Lange.
\newblock Unity: A platform for intelligent agents.
\newblock In \emph{CoRR}, 2018.

\bibitem[Marchesini et~al.(2023)Marchesini, Marzari, Farinelli, and Amato]{marchesini2023navigation}
E.~Marchesini, L.~Marzari, A.~Farinelli, and C.~Amato.
\newblock Safe deep reinforcement learning by verifying task-level properties.
\newblock In \emph{Proceedings of the 2023 International Conference on Autonomous Agents and Multiagent Systems}, page 1466–1475, 2023.

\end{thebibliography}

\clearpage
\onecolumn
\section*{Appendix}

\section{Proof of Lemma \ref{ourlemma}.}
\label{app:proof}
In order to show the monotonic improvement guarantees of \ourmethod, 
we want to bound the difference  $\vert \overline{\psi}(\pi') - \overline{L}_\pi(\pi')\vert$ where
\begin{equation*}
     \overline{\psi}(\pi') = (1-\varepsilon)\Bigg[\psi(\pi) + \mathbb{E}_{ \substack{ \tau \sim \pi' \\ s_0 \sim \rho}} \Big[\sum_{t=0}^\infty \gamma^t \Tilde{A}(s_t) \Big]\Bigg]+(\varepsilon)\Bigg[\psi(\pi) + \mathbb{E}_{ \substack{ \tau \sim \pi' \\ s_0 \sim \overline{\rho}}} \Big[\sum_{t=0}^\infty \gamma^t \Tilde{A}(s_t) \Big]\Bigg]
\end{equation*}

and 

\begin{equation*}
\overline{L}_\pi(\pi') = (1-\varepsilon)\Bigg[\psi(\pi) + \mathbb{E}_{\substack{ \tau \sim \pi \\ s_0 \sim \rho }} \Big[\sum_{t=0}^\infty \gamma^t \Tilde{A}(s_t) \Big]\Bigg] + (\varepsilon)\Bigg[\psi(\pi) + \mathbb{E}_{\substack{ \tau \sim \pi \\ s_0 \sim \overline{\rho}}} \Big[\sum_{t=0}^\infty \gamma^t \Tilde{A}(s_t) \Big]\Bigg].
\end{equation*}

Hence, following the proof of \citet{TRPO} we can derive a similar bound with simple algebra as:

\begin{align*}
    \footnotesize
    \vert \overline{\psi}(\pi') - \overline{L}_\pi(\pi')\vert = &(1-\varepsilon)\Bigg[\psi(\pi) + \mathbb{E}_{ \substack{ \tau \sim \pi' \\ s_0 \sim \rho}} \Big[\sum_{t=0}^\infty \gamma^t \Tilde{A}(s_t) \Big]\Bigg]+(\varepsilon)\Bigg[\psi(\pi) + \mathbb{E}_{ \substack{ \tau \sim \pi' \\ s_0 \sim \overline{\rho}}} \Big[\sum_{t=0}^\infty \gamma^t \Tilde{A}(s_t) \Big]\Bigg] -\\
    &(1-\varepsilon)\Bigg[\psi(\pi) + \mathbb{E}_{\substack{ \tau \sim \pi \\ s_0 \sim \rho }} \Big[\sum_{t=0}^\infty \gamma^t \Tilde{A}(s_t) \Big]\Bigg] + (\varepsilon)\Bigg[\psi(\pi) + \mathbb{E}_{\substack{ \tau \sim \pi \\ s_0 \sim \overline{\rho}}} \Big[\sum_{t=0}^\infty \gamma^t \Tilde{A}(s_t) \Big]\Bigg] \\
    = &(1-\varepsilon) \underbrace{\Bigg[\sum_{t=0}^\infty \gamma^t \Big\vert \mathbb{E}_{\substack{ \tau \sim \pi' \\ s_0 \sim \rho}} \Tilde{A}(s_t) -  \mathbb{E}_{\substack{ \tau \sim \pi \\ s_0 \sim \rho}} \Tilde{A}(s_t)\Big\vert \Bigg]}_{\leq \frac{4k\alpha^2\gamma}{(1-\gamma)^2} \text{by \citet{TRPO}}} + (\varepsilon) \Bigg[\sum_{t=0}^\infty \gamma^t \Big\vert \mathbb{E}_{\substack{ \tau \sim \pi' \\ s_0 \sim \overline{\rho}}} \Tilde{A}(s_t) -  \mathbb{E}_{\substack{ \tau \sim \pi \\ s_0 \sim \overline{\rho}}} \Tilde{A}(s_t)\Big\vert \Bigg] \\
    \leq &(1-\varepsilon)\Big[\frac{4k\alpha^2\gamma}{(1-\gamma)^2}\Big] + (\varepsilon) \Bigg[\sum_{t=0}^\infty \gamma^t \Big\vert \mathbb{E}_{\substack{ \tau \sim \pi' \\ s_0 \sim \overline{\rho}}} \Tilde{A}(s_t) -  \mathbb{E}_{\substack{ \tau \sim \pi \\ s_0 \sim \overline{\rho}}} \Tilde{A}(s_t)\Big\vert \Bigg].
\end{align*}

As stated in the main paper, from the fact that the $\alpha$-coupling definition (reported in Def. \ref{def:coupling}) is expressed over all possible states $s$, without any assumption or restriction on the initial state distribution, we have that 
$\pi$ and $\pi'$ are still $\alpha$-coupled even when we chose $s_0$ using a $\overline{\rho}$ as initial state distribution. 
Hence,

\begin{align*}
    \vert \overline{\psi}(\pi') - \overline{L}_\pi(\pi')\vert &\leq (1-\varepsilon)\Big[\frac{4k\alpha^2\gamma}{(1-\gamma)^2}\Big] + (\varepsilon)\Big[\frac{4k'\alpha^2\gamma}{(1-\gamma)^2}\Big]\\
    &= \frac{4\alpha^2\gamma}{(1-\gamma)^2}\Big[k + \varepsilon(k'-k)\Big] = \frac{4\alpha^2 \gamma}{(1-\gamma)^2}\Big[k(1-\varepsilon) + k'\varepsilon\Big].
\end{align*}

Next, note that since $0 \leq \varepsilon \leq 1$, we can derive that $k' \leq k$. In fact, we have:

\begin{align*}
 k(1-\varepsilon) + k'\varepsilon &\leq k\\
 -\varepsilon k + \epsilon k' &\leq 0\\
 k &\geq k'.
\end{align*}

Thus we conclude that 
$$
\frac{4\alpha^2 \gamma}{(1-\gamma)^2}\Big[k(1-\varepsilon) + k'\varepsilon\Big] \leq \frac{4\alpha^2 \gamma k}{(1-\gamma)^2}.
$$
The correctness of our result stems from the following facts: (i) $\alpha$-coupled policies are coupled over the entire $\mathcal{S}$, ensuring that policies are coupled even when combining $\rho$ and $\overline{\rho}$, and (ii) $k$ is defined as the maximum over all $(s, a) \in (\mathcal{S} \supseteq \mathcal{\overline{\mathcal{S}}}, \mathcal{A})$ and, as such, $k \geq k'$. In more detail, Lemma \ref{ourlemma} shows that enforcing the agent to start from retraining areas can help in narrowing the gap between the $\psi(\pi')$ and the local approximation $L_\pi(\pi')$. In fact, when $k=k'$, the difference reduces to the original case of Lemma \ref{lem:bound}. 
In contrast, when $k' < k$, the trajectories induced by $\overline{\rho}$ are not optimal since agents are penalized for violating the preference.
Thus, the agent gains a more precise understanding of the portion of the state space where the policy does not yield the maximum possible advantage. This motivates the design of \ourmethod\;that
linearly scales down $\varepsilon \to 0$, avoiding getting stuck in suboptimal restart distributions. $\quad\hfill \ensuremath{\Box}$

\section{Environments}
\label{subsec:envs}
We briefly describe the tasks and the desired behavioral specifications, referring to the original works for more details \citep{SafetyGymnasium, gymanm, marchesini2023navigation}. 

\textbf{Hopper, HalfCheetah (Figures \ref{fig:envs}a, b).} The robots have to learn how to run forward by exerting torques on the joints and observing the body parts' angles and velocities (for a total of 12 and 18 input features). The actions control the torques applied to the (3 and 6) joints of the robot. The agents are rewarded based on the distance between two consecutive time steps (i.e., positive and negative values for forward and backward movements). 
In these tasks, we have a velocity preference---agents should not go faster than $0.7402\frac{m}{s}$ and $3.2096\frac{m}{s}$, respectively. These are the same limitations considered in the safe RL literature \cite{ji2023omnisafe}, and the indicator cost deeming a velocity violation triggers when the limits are violated. The positive cost has a three-fold role:~it (i) starts the generation of a retrain area as described in Section \ref{sec:implementation_details}, using an initial size $\phi=0.01$ given by our initial grid search; (ii) triggers a small reward penalty to encourage the unconstrained baselines to avoid such an undesired behavior; and (iii) gets accumulated to model the constraints of the Lagrangian implementations. As in relevant literature, we set the constraints threshold to 25.

\textbf{Active Network Management (Figure \ref{fig:envs}c).} The agent has to reschedule the power generation of different renewable and fossil generators, to satisfy the energy demand of three loads connected to the power grid. Specifically, we observe the state of the power network through 18 features (i.e., active and reactive power injections, charge levels, and maximum productions). Moreover, we control power injections and curtailments using 6 continuous actions.
The behavioral preference here models energy losses and a grid's operational and transmission constraints. 
For this reason, we want our agent to limit penalties associated with violating these constraints. This task is significantly more complex than the previous ones since the agent is negatively rewarded based on the energy losses, the generation cost, and the violation of the operational constraints---however, to successfully solve the task in this particular environment, the agent \emph{must} receive penalties associated with energy losses, which are a natural consequence of power transmission. 
The indicator cost has the same role and consequences as the previous environments and is triggered when the agent violates the system's operational constraints. Based on the performance of the unconstrained baselines, we set the constraints threshold to 350. Due to the non-linear dynamics of power networks, a retrain area is generated using the exact violation state, with an initial bubble size $\omega=0$.

\textbf{Navigation (Figure \ref{fig:envs}d).} A mobile robot has to control its motor velocities to reach goals that randomly spawn in an obstacle-occluded environment without having a map. The agent observes the relative position of the goal and sparse lidar values sampled at a fixed angle (for a total of 22 features) and controls linear and angular velocity using 2 continuous actions.
Intuitively, the behavioral preference here models a safe behavior since we want the robot to avoid collisions. Similarly to the previous environments, the agent is positively (or negatively) rewarded based on its distance from the goal in two consecutive timesteps.
The indicator cost has the same consequences as the previous environments, and it is triggered upon every collision. We set the constraints threshold to 20, indicating the robots should not collide for more than 20 steps in a training episode that lasts for 500 steps. We consider the simulated lidar precision to initialize the bubble size $\phi=0.025$.

\section{Missing Plots from Section \ref{sec:experiments}}
\label{suppl:results}
\begin{figure}[h!]
    \centering
    \includegraphics[width=\linewidth]{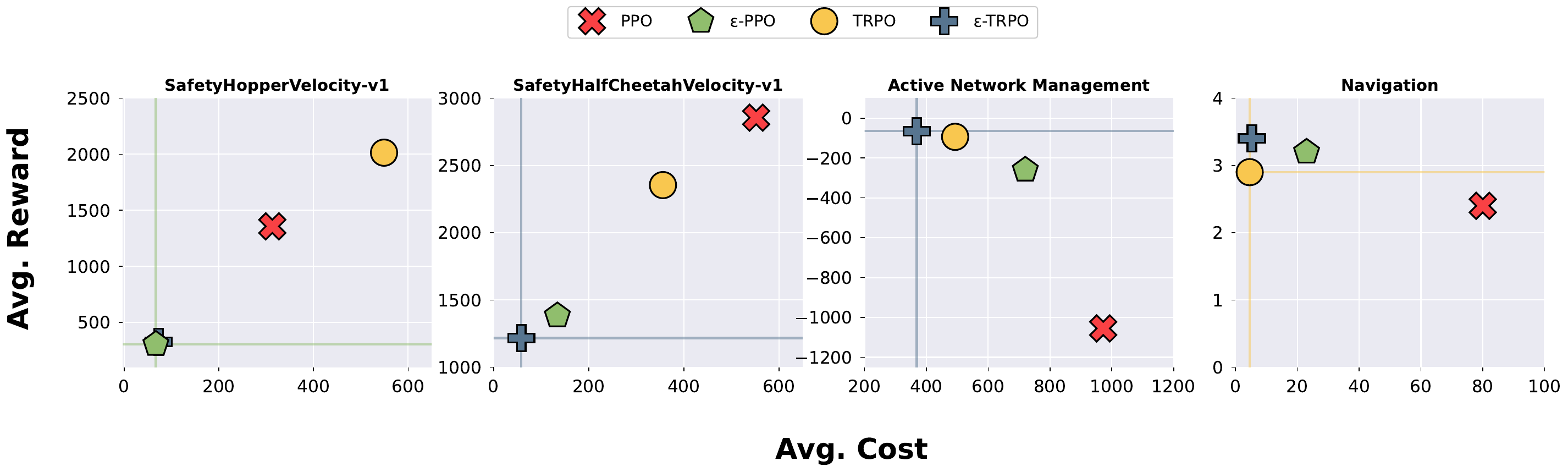}
    \caption{Pareto frontier of $\varepsilon$-PPO, $\varepsilon$-TRPO, PPO and TRPO at convergence in four different environments.  Each column (i.e., each task) shows the average reward and cost during the training. Learning curves are reported in Figure \ref{fig:base_results} in the main paper.}
    \label{fig:base_results_pareto}
\end{figure}

 Figure \ref{fig:base_results_pareto} shows the Pareto frontier reporting on the y-axis the average reward and on the x-axis the average cost at convergence. \ourmethod\;in general allows to achieve the best trade-off between average reward and cost (no other methods reach better performance in the upper-left corner), i.e., less violation of the behavioral desiderata while still successfully solving the task. It is important to notice that only in the navigation scenario the cost value of TRPO at convergence is slightly lower than $\varepsilon-$TRPO one. However, as reported in the learning curves in the main papers, our approach results in significantly more sample efficiency than the baseline counterparts.

\begin{figure}[h!]
    \centering
    \includegraphics[width=\linewidth]{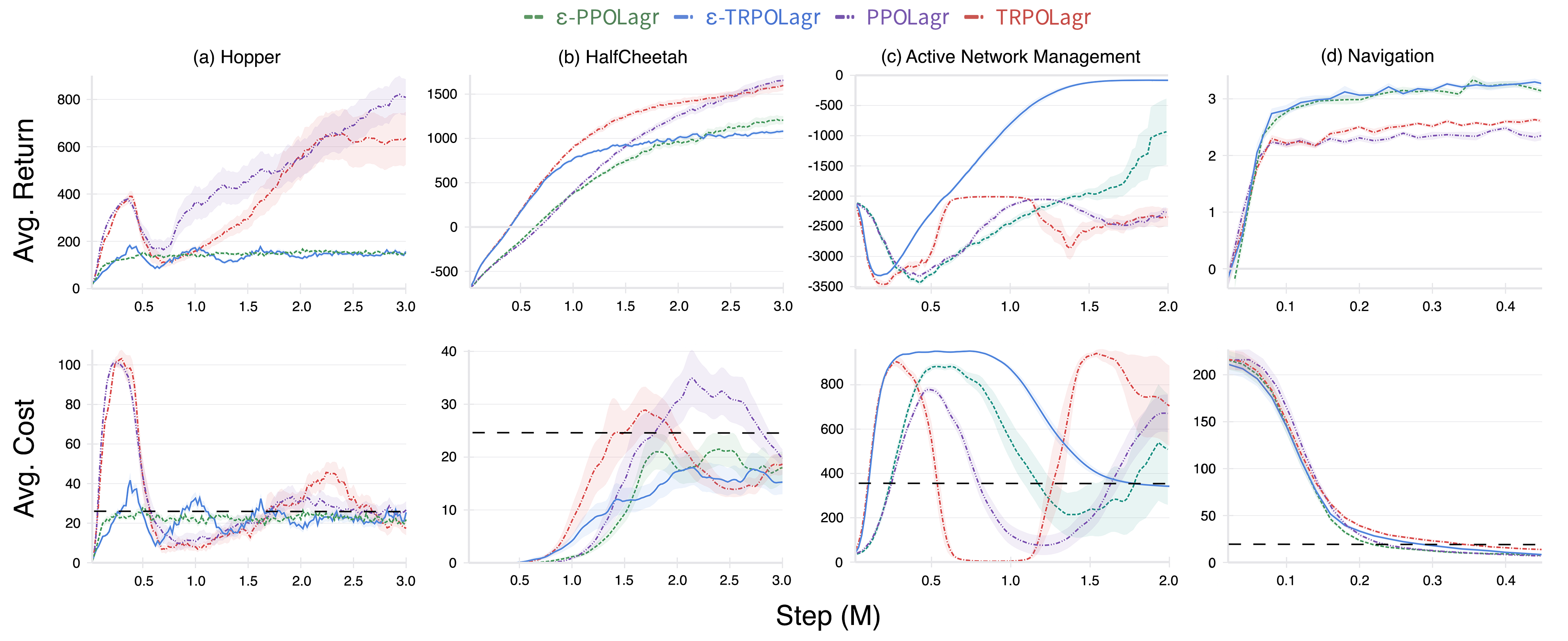}
    \caption{Comparison of $\varepsilon$-PPOLagr, $\varepsilon$-TRPOLagr, PPOLagr and TRPOLagr in four different environments. Each column (i.e., each task) shows the average reward and cost during the training. The black dotted line represents the cost threshold.}
    \label{fig:lag_results}
\end{figure}

In Figure \ref{fig:lag_results}, we report the training performance of original constrained policy optimization algorithms and the one enhanced with \ourmethod. In detail, our approach allows us to reduce the constraint violations and, in the more complex tasks, allow agents to improve performance.
\newpage
\section{Environmental Impact}
\label{app:env_impact}
Despite each individual training run being ``relatively'' computationally inexpensive due to the use of CPUs, the $\approx1600$ experiments of our evaluation led to cumulative environmental impacts due to computations that run on computer clusters for an extended time. Our experiments were conducted using a private infrastructure with a carbon efficiency of $\approx 0.275 \frac{\text{kgCO$_2$eq}}{\text{kWh}}$, requiring a cumulative $\approx$240 hours of computation. Total emissions are estimated to be $\approx6.93 \text{kgCO$_2$eq}$ using the \href{https://mlco2.github.io/impact#compute}{Machine Learning Impact calculator}, and we purchased offsets for this amount through \href{https://www.treedom.net}{Treedom}.

\section{Hyper-parameters}
\label{app:hyperparameters}
Regarding the baselines, we performed an initial grid search, but the original parameters of the omnisafe library resulted in the best performance \citep{ji2023omnisafe}.
Table \ref{tab:gridsearch} lists the key hyper-parameters considered in our initial grid search for TRPO, PPO, their Lagrangian versions, and \ourmethod. The best parameters used in our evaluation are highlighted in the last column. 
\begin{table}[h]
\centering
\caption{Hyper-parameters candidate for initial grid search tuning, and best parameters.}
\label{tab:gridsearch}
\begin{tabular}{llll}
\toprule
         & \textbf{Parameter}           & \textbf{Grid Search}           & \textbf{Best Values}  \\ \midrule
\textbf{Policy Optimization}    
    & Steps per epoch       & 20000, 30000    & 20000 \\

    & Update iterations       & 10, 20    & 10 \\
    & Batch size       & 64, 128    & 128 \\
    & Target KL       & 0.01, 0.001    & 0.01 \\
    & Max grad. norm       & 20, 40    & 40 \\
    & $\gamma$            & 0.9, 0.95, 0.99 & 0.99    \\
    & GAE       & 0.95            & 0.95   \\
    & PPO clip       & 0.2            & 0.2   \\
    & Penalty       & 0.0, 0.1, 0.25            & 0.1   \\ \midrule
\textbf{Actor-Critic networks}  
    & N° layers           & 2                   & 2   \\
    & Size                & 64, 128             & 64 \\ 
    & Activation          & tanh                & tanh \\
    & Optimizer           & Adam                & Adam \\
    & Learning rate       & 1e-3, 3e-4, 5e-5    & 3e-4   \\    \midrule
\textbf{Lagrangian}  
    & Multiplier init.    & 0.001      & 0.001   \\
    & Multiplier learning rate & 0.035, 0.0035  & 0.035 \\ 
    & Optimizer           & Adam                & Adam \\
    & Learning rate       & 1e-3, 3e-4, 5e-5    & 3e-4   \\    
    & Cost limits         & -                & Section \ref{sec:implementation_details} \\
    & Cost $\gamma$         & 0.99            0.99 \\
    & Cost GAE    & 0.95            & 0.95   \\
\midrule
    
\textbf{\ourmethod}  
    & Bubble size init $\omega$ & 0.0, 0.01, 0.025                 & 0.0, 0.01, 0.025   \\
    & Similarity $\beta$ & 0.01, 0.03 & 0.03  \\
    & Max retrain areas  & 250, 500, 1000 & 500   \\
    & $\varepsilon$-decay  & 0.25, 0.5, 0.75 & 0.75  \\
    & Minimum $\varepsilon$  & 0.0, 0.25, 0.5 & 0.5 
    \\ \bottomrule
\end{tabular}
\end{table}

\end{document}